# Optimal Rectangle Packing:
# An Absolute Placement Approach

**Eric Huang**                                                    EHUANG@PARC.COM
*Palo Alto Research Center*
*3333 Coyote Hill Road*
*Palo Alto, CA 94304 USA*

**Richard E. Korf**                                               KORF@CS.UCLA.EDU
*UCLA Computer Science Department*
*4532E Boelter Hall*
*University of California, Los Angeles*
*Los Angeles, CA 90095-1596 USA*

## Abstract

We consider the problem of finding all enclosing rectangles of minimum area that can contain a given set of rectangles without overlap. Our rectangle packer chooses the $x$-coordinates of all the rectangles before any of the $y$-coordinates. We then transform the problem into a perfect-packing problem with no empty space by adding additional rectangles. To determine the $y$-coordinates, we branch on the different rectangles that can be placed in each empty position. Our packer allows us to extend the known solutions for a consecutive-square benchmark from 27 to 32 squares. We also introduce three new benchmarks, avoiding properties that make a benchmark easy, such as rectangles with shared dimensions. Our third benchmark consists of rectangles of increasingly high precision. To pack them efficiently, we limit the rectangles' coordinates and the bounding box dimensions to the set of subset sums of the rectangles' dimensions. Overall, our algorithms represent the current state-of-the-art for this problem, outperforming other algorithms by orders of magnitude, depending on the benchmark.

## 1. Introduction

Given a set of rectangles, our problem is to find all enclosing rectangles of minimum area that will contain them without overlap. We refer to an enclosing rectangle as a *bounding box*, to avoid confusion. The optimization problem is NP-hard, while the problem of deciding whether a set of rectangles can be packed in a given bounding box is NP-complete, via a reduction from bin-packing (Korf, 2003). The *consecutive-square benchmark* is a simple set of increasingly difficult benchmarks for this problem, where the task is to find the bounding boxes of minimum area that contain a set of squares of dimensions $1 \times 1$, $2 \times 2$, ..., up to $N \times N$ (Korf, 2003). For example, Figure 1 is an optimal solution for $N$=32. We will use this benchmark to explain many of the ideas in this paper, but our techniques are not limited to packing squares, and apply to all rectangles.

Rectangle packing has many practical applications, including modeling some scheduling problems where tasks require resources that are allocated in contiguous chunks. For example, consider the task of scheduling and allocating contiguous memory addresses to programs. The width of a rectangle represents the length of time a program runs, and the height represents the amount of contiguous memory it needs. A rectangle packing solution tells us both when programs should be run, as well as which memory addresses they should be assigned. Similar problems include scheduling when and where ships of different length can be berthed along a single, long wharf (Li, Leong, & Quek, 2004), as well as the allocation and scheduling of radio frequency spectra usage (Mitola & Maguire, 1999). Rectangle packing also appears when loading a set of rectangular objects on a pallet without stacking them. Some cutting stock and layout problems also contain rectangle packing subproblems.

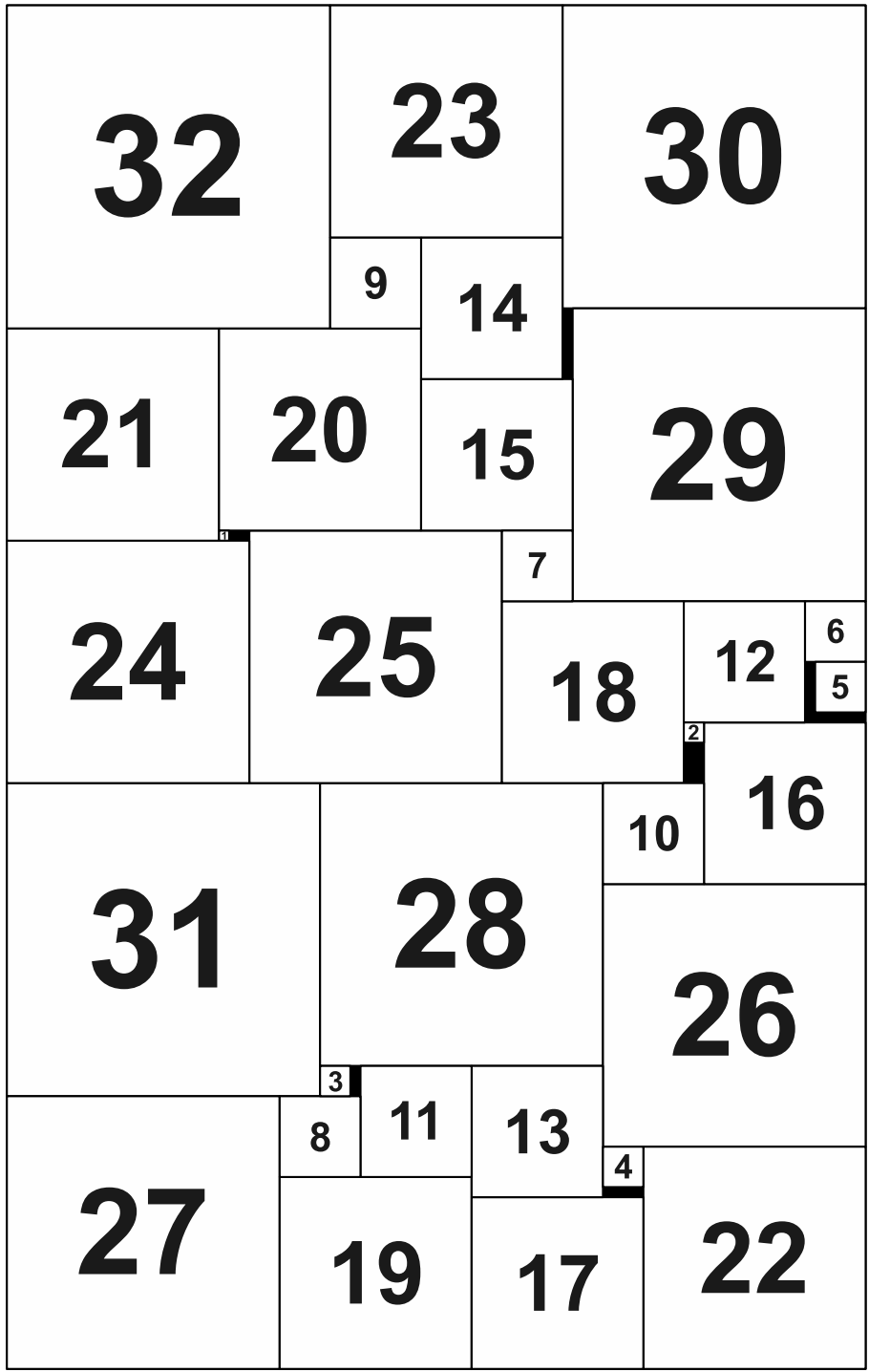

Figure 1: An optimal solution for $N$=32 of the consecutive-square benchmark, packing squares of dimensions $1 \times 1$, $2 \times 2$, ..., $31 \times 31$, and $32 \times 32$ in a bounding box of minimum area, which is $85 \times 135$.

## 1.1 Overview

The remainder of this article is organized as follows. We first introduce various benchmarks in Section 2 that specifically define the rectangle packing instances we will solve. In Section 3, we review the state-of-the-art rectangle packers and their techniques, which provides a foundation upon which we present our new work. We follow in Section 4 with the data collected and compare our work against the previous state-of-the-art using previous benchmarks. We also compare the difficulty of previous benchmarks with our new ones.

In Section 5, we present a benchmark of rectangles of successively higher precision dimensions, new solution techniques to handle this, and follow with experimental results. Then we compare our methods to the competing search spaces used for packing high-precision rectangles, and show that our methods remain competitive.

Sections 6 and 7 explain various avenues for future work, concluding this article by summarizing all of our contributions and results. We have previously published much of this work in several conference papers (Huang & Korf, 2009, 2010, 2011).

## 2. Benchmarks

There are several reasons motivating our benchmarks. First, our benchmarks describe instances with a single parameter $N$, allowing researchers to easily reproduce the instances. Second, because the instances are unique, optimal solutions that are reported can be easily validated by others. These are advantages over many real-world instance libraries and randomly generated ones. Third, our benchmarks define an infinite set of instances where each successive instance is harder than the previous. A solver is superior to another solver if it can solve the same instance faster, or a larger instance in the same amount of time. By contrast, comparison using a library of instances may require counting the number of instances that are completed within a given time limit. Furthermore, with instance libraries, often one solver performs well on one subset of instances while a competing solver performs well on a different subset, making such comparisons inconclusive.

We believe our benchmarks capture some of the more difficult instances a rectangle packer may face so we do not investigate the modeling and generation of random problems. Although Clautiaux et al. (2007) and others have used random instances, the non-random benchmarks used by Korf (2003) and Simonis and O'Sullivan (2008) have better facilitated the comparison of state-of-the-art packers. However, for more comprehensive overviews, we refer the reader to the numerous surveys available (Lodi, Martello, & Vigo, 2002; Lodi, Martello, & Monaci, 2002; Dowsland & Dowsland, 1992; Sweeney & Paternoster, 1992).

## 2.1 Previous Benchmarks

Several of the previous benchmarks used in the literature can be shown to be easier than the benchmarks that we propose. Part of this is due to the fact that benchmarks, like solvers, may also be improved with further research, to ensure that they cover various properties of rectangles, in addition to providing an easy way to compare performance among different packers and measure progress.

The *consecutive-square benchmark* (Korf, 2003), is a simple set of increasingly difficult instances, where the task is to find all bounding boxes of minimum area that contain a set of squares of sizes $1 \times 1$, $2 \times 2$, ..., up to $N \times N$. Prior to our work, many of the recent state-of-the-art packers used this popular benchmark to measure performance, including that of Moffitt and Pollack (2006), Korf, Moffitt, and Pollack (2010), and Simonis and O'Sullivan (2008). To date, the largest instance solved for this problem is $N=32$, shown in Figure 1, using our packer (Huang & Korf, 2009). We do not consider the problem of packing squares in a square because this benchmark gets much easier as the problem size increases, due to large differences in the areas of consecutive square bounding boxes.

In the *unoriented consecutive-rectangle benchmark* (Korf et al., 2010), an instance is a set of rectangles of sizes $1 \times 2$, $2 \times 3$, ..., up to $N \times (N+1)$, and rectangles may be rotated by 90-degrees. As we will subsequently explain, the fact that there are many pairs of rectangles in this instance which share equal dimensions causes the optimal solutions to leave no empty space, making this benchmark easy to solve. We include this benchmark for completeness, but note that it is not an effective measure for comparing different packers.

*Finding only the first optimal solution* is another benchmark Simonis and O'Sullivan (2011) have used in conjunction with problem instances from the unoriented consecutive-rectangle benchmark. In contrast to our problem of finding all optimal solutions, they measure the time it takes to find only the first optimal solution, which makes it much more difficult to reliably compare against other solvers unless the focus of the research is on value ordering and tie-breaking among bounding boxes of equal area.

For example, Simonis and O'Sullivan (2011) report that to find the first solution to $N=26$ takes 3:28:20 (3 hours, 28 minutes, and 20 seconds). As shown in Table 8 on page 72, there are six solutions for $N=26$: $42 \times 156$, $52 \times 126$, $56 \times 117$, $63 \times 104$, $72 \times 91$, $78 \times 84$, each requiring our solver CPU times of 0:32, 41:40, 53:19, 1:55:04, 1:33:22, and 8:53:01, respectively. There are no smaller bounding boxes we needed to test because the optimal solution has no empty space, so if we used Simonis and O'Sullivan's termination criteria and just returned the first optimal solution, we would only need 32 seconds. Therefore, finding all minimum bounding boxes instead of just the first one is a benchmark which produces harder problems for larger $N$, and better facilitates program comparisons.

## 2.2 Properties of Easy Benchmarks to Avoid

To motivate our new benchmarks, we will now explain why the previous benchmarks tended to be much easier in comparison, and why we have constructed our new benchmarks to describe instances consisting of rectangles with unique dimensions, without duplicates, and without most of the area being occupied by only a few rectangles.

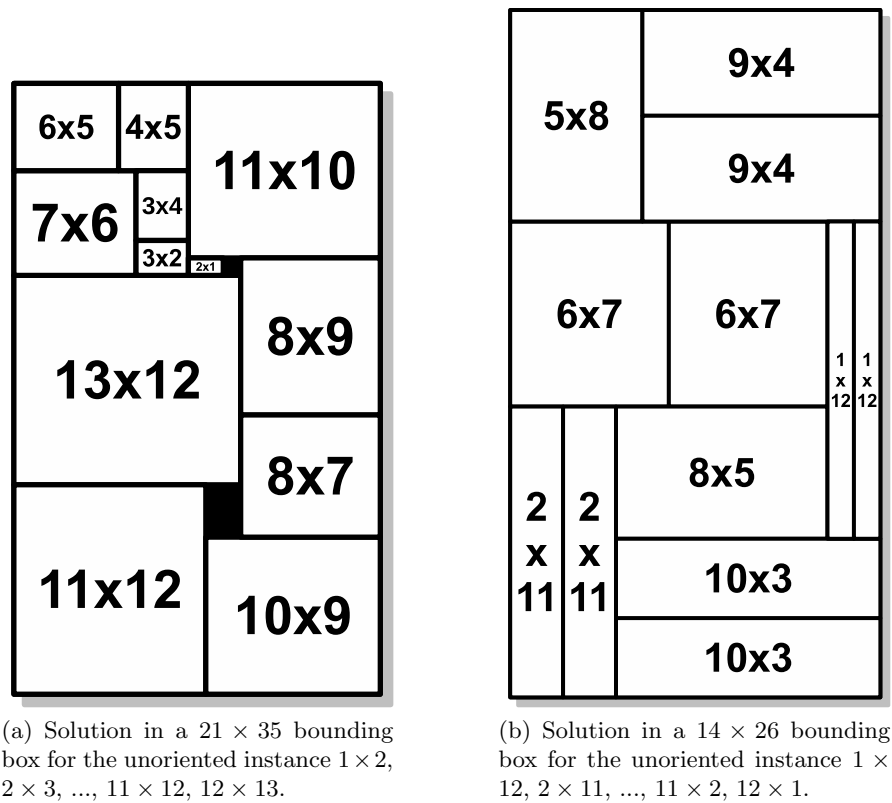

(a) Solution in a $21 \times 35$ bounding box for the unoriented instance $1 \times 2$, $2 \times 3$, ..., $11 \times 12$, $12 \times 13$.

(b) Solution in a $14 \times 26$ bounding box for the unoriented instance $1 \times 12$, $2 \times 11$, ..., $11 \times 2$, $12 \times 1$.

Figure 2: Examples of solutions for instances of rectangles with equal dimensions.

### 2.2.1 Rectangles With Equal Dimensions

In the unoriented consecutive-rectangle benchmark, all rectangles share a dimension with another rectangle. For example, Figure 2a is an optimal solution for $N$=12. In optimal solutions, rectangles of equal dimensions tend to line up next to each other, forming larger rectangles and leaving little empty space. In Figure 2a, the $8 \times 9$ and $7 \times 8$ line up, as do the $5 \times 6$ with the $4 \times 5$, and the $3 \times 4$ with the $2 \times 3$. In fact, the solutions to this benchmark all have a much smaller percentage of empty space than similar-sized instances from the consecutive-square benchmark, where all rectangles have unique dimensions. We also notice that benchmarks with duplicate rectangles, such as that in Figure 2b, are solved quickly.

### 2.2.2 Rectangles With Small Area and Small Dimensions

Figure 2b is also an example of a *perfect packing*, because there is no empty space in the solution. Problems with perfect packings tend to be easy for two reasons. One is that if we test bounding boxes in increasing order of area, we test fewer boxes, since we never test a box with more than the minimum area required. The second is that for these problems, rather than deciding for each rectangle where it should go in the bounding box, a more efficient algorithm is to decide for each cell of empty space which rectangle should occupy it. As soon as a small region of empty space is created that can't accomodate any remaining rectangles, the algorithm can backtrack.

In both the consecutive-square and the unoriented rectangle benchmarks, a few large rectangles capture much of the total area in an instance. Thus, the packer does not search too deeply before using up the allowable empty space. With little empty space, early backtracking is very likely since we cannot find a place for the next rectangle. Therefore, small rectangles in these benchmarks have an insignificant impact on the search effort.

In previous benchmarks, such as the consecutive-square benchmark, the retangles with the largest area also have the largest dimensions, making it obvious which rectangles to place first, because the largest rectangles are the most constrained, and impose the most constraints on the remaining rectangles.

By contrast, in our new benchmarks there is a trade-off between rectangles with large dimensions and those with large area. The widest rectangle in our oriented equal-perimeter benchmark, described below, has the smallest branching factor as we search for $x$-coordinates. However, it also has the least area, so during search it won't constrain the placement of the remaining rectangles much. This raises the non-trivial question of the best variable ordering for non-square rectangles.

## 2.3 New Benchmarks

We propose several new benchmarks that are more difficult when comparing instances with the same number of rectangles. Our experimental results make use of the following benchmarks, in addition to the consecutive-square and unoriented consecutive-rectangle benchmarks described above.

### 2.3.1 EQUAL-PERIMETER RECTANGLES

First, we present the *oriented equal-perimeter rectangle benchmark*, where each instance is a set of rectangles of sizes $1 \times N$, $2 \times (N-1)$, ..., $(N-1) \times 2$, $N \times 1$, and rectangles may not be rotated (see Figure 3). Given $N$, all rectangles are unique and have a perimeter of $2N+2$. In our experiments, this benchmark is much more difficult than either the consecutive-square benchmark or the unoriented consecutive-rectangle benchmark (Korf et al., 2010) for the same number of rectangles. We tested our state-of-the-art packer (Huang & Korf, 2010) on both old and new benchmarks. $N=22$ from our oriented equal-perimeter benchmark took over nine hours to solve, while $N=22$ from the consecutive-square and unoriented consecutive-rectangle benchmarks took only one second and six seconds, respectively.

Second, we present the *unoriented double-perimeter rectangle benchmark*, where instances are described as a set of rectangles $1 \times (2N-1)$, $2 \times (2N-2)$, ..., $(N-1) \times (N+1)$, $N \times N$, and rectangles may be rotated by 90-degrees. All rectangles here are unique and have a perimeter of $4N$. Not only is this benchmark more difficult than the benchmarks used previously in the literature, but this benchmark also is more difficult than the oriented one we introduced in the previous paragraph. In our experiments using all of our techniques, $N=18$ took over two days to solve.

So far, the benchmarks that we have discussed all have low-precision integer dimensions. This property poses no problem for our packer, which enumerates the various integer coordinate locations where a rectangle may be placed. With high-precision values, however, the number of distinct positions increases dramatically. This motivates our study of packing rectangles with high-precision dimensions. In particular, we propose the *unoriented high-precision rectangle benchmark*, where instances are described as a set of rectangles $\frac{1}{1} \times \frac{1}{2}, \frac{1}{2} \times \frac{1}{3}, ..., $ up to $\frac{1}{N} \times \frac{1}{N+1}$. The methods used to solve this benchmark are quite different from those used in the low-precision case.

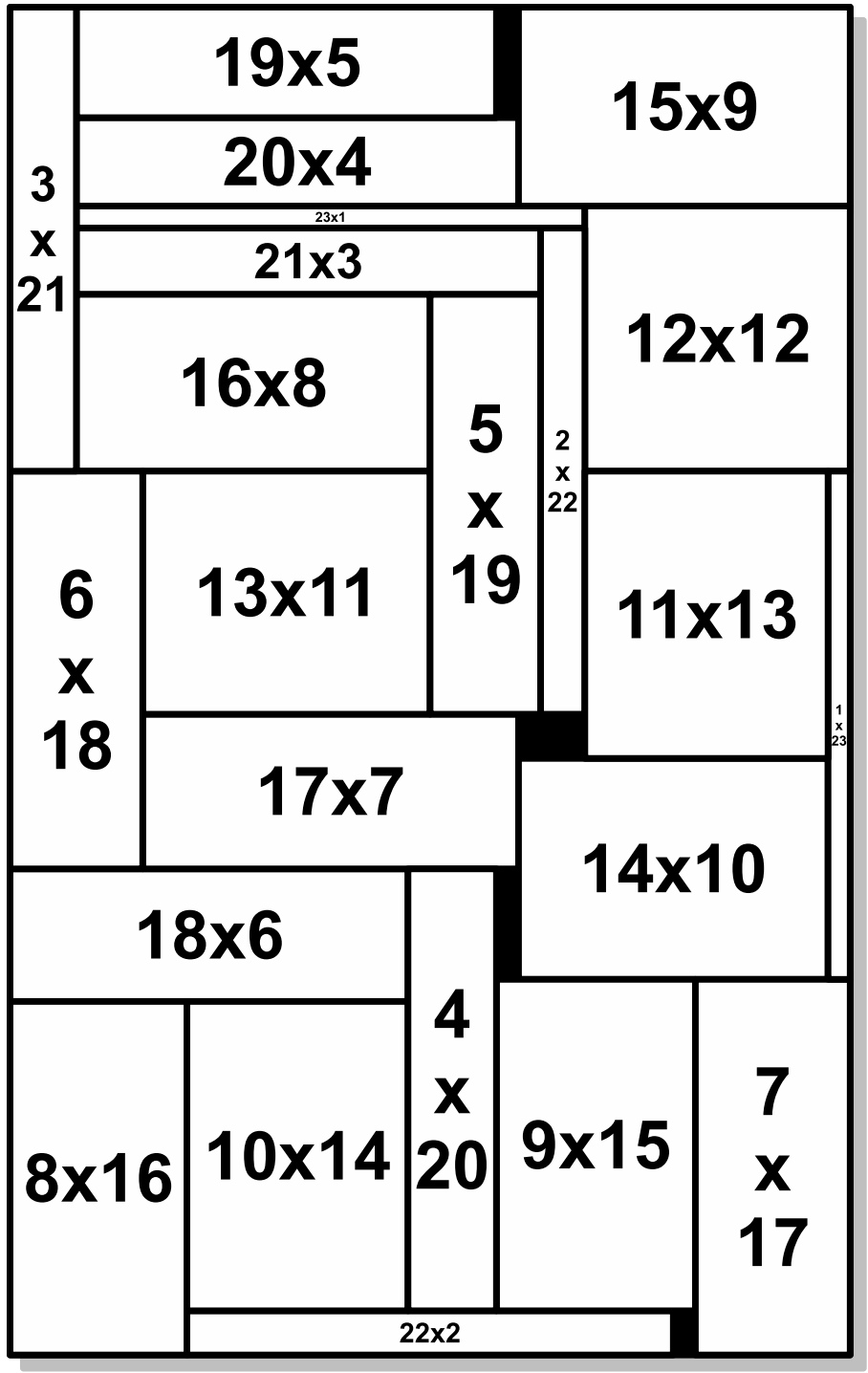

Figure 3: An optimal solution for $N$=23 of the oriented equal-perimeter benchmark, packing oriented rectangles of dimensions $1 \times 23$, $2 \times 22$, ..., $22 \times 2$, and $23 \times 1$ in a bounding box of minimum area, which is $38 \times 61$.

## 3. Solution Techniques

In this section we describe previous solution strategies as well as the various new techniques we use in our rectangle packer. We first describe our techniques as they apply to the consecutive-square benchmark, the oriented equal-perimeter benchmark, and the unoriented double-perimeter benchmark. Our work on the unoriented high-precision rectangle benchmark is not included here because the methods are significantly different, and is deferred to Section 5.

### 3.1 Previous Work

Some of the earlier work that focused on optimal methods for packing a set of rectangles in a given bounding box were motivated by the problem of pallet loading. Dowsland (1987) used depth-first search on an abstract graph representation of the search space to solve the problem optimally on problem sets modeled after real-world pallet and box dimensions. Although her problem instances contained an average of 30 rectangles and up to 50, her benchmarks were far easier than those we consider here, as all of the rectangles were the same size, and there was a significant amount of empty space in the solutions. Bhattacharya et al. (1998) extended the work with additional lower bounds and pruning techniques based on dominance conditions and demonstrated their work on the same benchmarks.

In examining rectangle packing instances where rectangles are of different dimensions, Onodera et al. (1991) used depth-first search, in which each branching point in their search space was a commitment to a particular non-overlap constraint between two rectangles. Lower bound and graph reduction techniques were applied to prune the search space, allowing them to optimally solve problems with up to six rectangles.

Chan and Markov's BloBB (2004) packer used branch-and-bound in order to find the minimum area bounding box that can contain a set of rectangles. Their solver could handle up to eleven rectangles, and they observed that instances with duplicate rectangles were much easier, causing their packer to cluster such rectangles together in an optimal solution. Lesh et al.'s solver (2004) used depth-first search, placing each rectangle first in the bottommost and left-most position in which it fit (the bottom-left heuristic, see Chazelle, 1983), to determine whether or not a set of rectangles can be packed in a given enclosing rectangle. They were able to handle about twenty-nine rectangles in ten minutes on average, but their testbed consisted only of instances whose optimal solutions had no empty space.

Clautiaux et al. (2007) presented a branch-and-bound method in which all the $x$-coordinates for the rectangles were computed prior to any of the $y$-coordinates. While assigning $x$-coordinates, their method uses a relaxation similar to the *cumulative constraint* (Aggoun & Beldiceanu, 1993) which requires that the sum of the heights of all rectangles overlapping a particular $x$-coordinate cannot exceed the height of the bounding box. The $y$-coordinates are then determined using a search space derived from the bottom-left heuristic (Chazelle, 1983), using optimized data structures from Martello and Vigo (1998). Beldiceanu and Carlsson (2001) applied the plane sweep algorithm used in computational geometry to detect violations of the non-overlap constraints, and later adapted the technique to a geometric constraint kernel (Beldiceanu, Carlsson, Poder, Sadek, & Truchet, 2007). Lipovetskii (2008) proposed a branch-and-bound algorithm that placed rectangles in the lower-left hand positions.

The prior state-of-the-art, due to Korf (2003, 2004) and Simonis and O'Sullivan (2008), both divide the rectangle packing problem into the *containment problem* and the *minimal bounding box problem*. The former tries to pack a given set of rectangles in a given bounding box, while the latter finds the bounding box of least area that can contain the given set of rectangles. In both packers the algorithm for the minimal bounding box problem calls the algorithm for the containment problem as a subroutine.

## 3.2 Our Overall Search Strategy

Like Korf et al.'s (2010) algorithm, we have a minimum bounding box solver which calls a containment problem solver, and like Simonis and O'Sullivan (2008), we assign $x$-coordinates prior to any of the $y$-coordinates.

Although we use some of Simonis and O'Sullivan's (2008) ideas, we do not take a constraint programming approach in which all constraints are specified to a general-purpose solver like Prolog. Instead, we implemented our program from scratch in C++, allowing us to more flexibly choose which constraints to use at what time and to naturally encode the search space we use for the $y$-coordinates. We implemented a chronological backtracking algorithm with dynamic variable ordering. Our algorithm works in five stages as it goes from the root of the search tree down to the leaves:

1. The minimum bounding box algorithm generates an initial candidate set of bounding boxes of various widths and heights.

2. The containment solver is called for each bounding box in order of increasing area, and for each infeasible bounding box, we insert another back into the candidate set of bounding boxes with a height one unit greater. If a packing was found, we continue testing boxes of equal area to find all optimal solutions before terminating.

3. The containment solver first works on the $x$-coordinates in a model where variables are rectangles and values are $x$-coordinate locations, using dynamic variable ordering and a constraint that detects infeasible subtrees.

4. For each $x$-coordinate solution found, the problem is transformed into a perfect packing instance.

5. It then searches for a set of $y$-coordinates in a model where variables are empty corners and values are rectangles.

We now describe in detail each of these steps.

### 3.3 Minimum Bounding Box Problem

One way to solve the minimum bounding box problem is to find the minimum and maximum areas describing the set of candidate and potentially optimal bounding boxes. Boxes of all sizes are generated with areas within this range, and then tested in non-decreasing order of area until all solutions of smallest area are found. A lower bound on the area is the sum of the areas of the given rectangles. An upper bound on the area is determined by the bounding box of a greedy solution found by setting the bounding box height to that of the tallest rectangle, and then placing the rectangles in the first available position when scanning from left to right, and for each column scanning from bottom to top.

There are several techniques (Korf, 2003, 2004) that we use to prune the set of bounding boxes, which we review here. We first generate a set of widths for our bounding boxes, starting with the width of the widest rectangle up to the width of the greedy solution described above. Then for each width, we generate a feasible height using lower bounds which we will subsequently describe. The resulting bounding boxes are used to initialize a min-heap sorted in non-decreasing order of area. The search proceeds by calling the containment solver on the bounding box of minimal area in this heap. If the box is infeasible, then we increase the height of the box by one, and insert the new box back into the min-heap.

For a given bounding box width, we initialize its height to the maximum of the following lower bounds. First, the height must be at least the height of the tallest rectangle in the instance. Second, the height must be large enough to accommodate the total area of the rectangles in the instance. Third, for every pair of rectangles, if the sum of their widths exceed the width of the bounding box, then the bounding box height must be at least the sum of their heights, since they can't appear side-by-side, but one must be on top of the other. Fourth, the set of rectangles whose widths are greater than half the width of the bounding box must all be stacked vertically, including the rectangle of smallest height whose width is exactly half the width of the bounding box. Finally, if certain properties exist for a given rectangle packing instance, we force the height to be greater than or equal to the width to break symmetry. For example, one sufficient property is having an instance consisting of just squares, since a solution in a $W \times H$ bounding box easily transforms into another one in a $H \times W$ bounding box. Another sufficient property is when every rectangle of dimensions $w \times h$ can correspond to another one of dimensions $h \times w$.

For unoriented instances, given a bounding box width, certain rectangles may be forced into one orientation, improving the lower bound on the bounding box height. Note that we can also break the symmetry on the bounding box dimensions for every unoriented instance.

### 3.3.1 ANYTIME ALGORITHM

In a problem instance with many rectangles, or when an immediate solution is required, Korf (2003) provides an anytime algorithm for the bounding box problem, replacing the one described above, which also calls the containment problem solver. We first find a greedy solution on a bounding box whose height is equal to the tallest rectangle, as described in the previous section. We then repeatedly call the containment problem solver in the following way. If the previous attempt for a given bounding box resulted in a packing or if its area is greater than the area of the best solution seen so far, then we decrease the width by one unit and attempt to solve the resulting bounding box problem. If instead the previous attempt were infeasible, then we increase the height of the bounding box by one unit. The algorithm terminates when the width of the current bounding box is less than the width of the widest rectangle.

### 3.4 Containment Problem

Korf's (2003) absolute placement approach modeled rectangles as variables and positions in the bounding box as values. Rectangles were placed in turn with a depth-first search, and all possible locations were tested for each rectangle. By contrast, Simonis and O'Sullivan's (2008) packer assigned the $x$-coordinates of all the rectangles before any of the $y$-coordinates, as suggested by Clautiaux et al. (2007), as well as using the cumulative constraint (Aggoun & Beldiceanu, 1993), improving performance by orders of magnitude. The cumulative constraint adds the height of all the rectangles that overlap a given $x$-coordinate location, pruning if any of these values exceed the height of the bounding box. This constraint was checked while exploring $x$-coordinates and also while exploring $y$-coordinates later on. We improved on this by exploring the $y$-coordinates differently, modeling candidate locations as variables, and rectangles as values (Huang & Korf, 2009), which made our packer over an order of magnitude faster than that of Simonis and O'Sullivan's.

Simonis and O'Sullivan (2008) furthermore applied the *least-commitment principle* (Yap, 2004) from constraint processing, by first committing the placement of rectangles to an interval of $x$-coordinates instead of just a single $x$-coordinate value. These $x$-intervals are explored in turn, and constrain the candidate individual $x$-coordinates explored later. This works because committing to an $x$-interval can induce pruning via the cumulative constraint. For example, picking an $x$-interval of $[a, b]$ with a size that is smaller than the width of the rectangle $w_r$, implies that regardless of which $x$-coordinate the rectangle eventually takes, it must contribute its height to each $x$-coordinate within the interval $[b, a + w_r]$. Finally, the height of the bounding box constrains the cumulative heights of all rectangles for any given $x$-coordinate, similar to the ideas of Beldiceanu et al. (2008). Larger intervals result in weaker constraint propagation (less pruning) but a smaller branching factor, while smaller intervals result in stronger constraint propagation but a larger branching factor. The size of the intervals are experimentally determined.

For example, a $4 \times 2$ rectangle with $x$-coordinates restricted to the interval [0,2] contributes a height of 2 at $x$-coordinates 2 and 3 even prior to deciding its exact $x$-coordinate value. This *compulsory part* (Lahrichi, 1982) constrains the cumulative height of the rectangles that may overlap $x$-coordinates 2 and 3 in the solution. If these interval assignments were all infeasible, then searching for individual $x$-values is futile. However, if we do find a set of interval assignments, then we still have to search for a set of single $x$-coordinate values. Simonis and O'Sullivan (2008) assigned $x$-intervals, single $x$-coordinates, $y$-intervals, and single $y$-coordinates, in that order.

### 3.5 Assigning $X$-Intervals and $X$-Coordinates

For the $x$-coordinates, we propose a pruning constraint adapted from Korf's (2003) wasted-space pruning heuristic, a dynamic variable order to replace Beldiceanu's (2008) fixed ordering, and a method to optimize the values assigned to our $x$-interval variables.

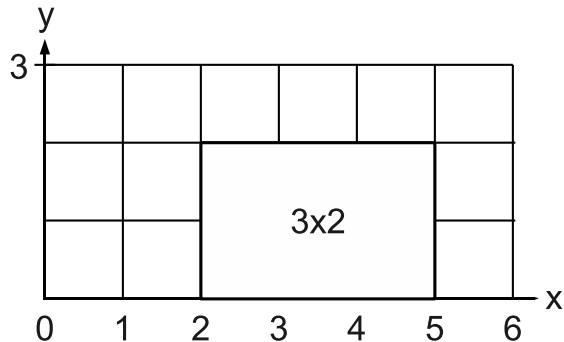

Figure 4: To test for violations of the cumulative constraint, the remaining space after placing a $3 \times 2$ rectangle at $x{=}2$ is represented as the vector $\langle 3, 3, 1, 1, 1, 3 \rangle$.

### 3.5.1 PRUNING INFEASIBLE SUBTREES

We present a constraint-based formulation of Korf's (2003) two-dimensional wasted space pruning algorithm, adapted to the one-dimensional case. Given a partial solution, Korf's algorithm computed a lower bound on the amount of wasted space, which was then used to prune against an upper bound. By contrast, we do not compute any numerical bounds and instead detect infeasibility with a single constraint.

As rectangles are placed in the bounding box, the remaining empty space gets chopped up into small irregular regions. Eventually the empty space is segmented into small enough chunks such that they cannot accommodate any of the remaining unplaced rectangles, at which point we backtrack. While assigning $x$-coordinates in a bounding box of height $H$, we keep a histogram $\langle v_1, v_2, \ldots, v_H \rangle$, where $v_i$ is the number of empty cells (units of empty space) that are in empty columns of height $i$. For example, assume that in Figure 4 we assigned only the $x$-coordinates of a $3 \times 2$ rectangle in a $6 \times 3$ bounding box. The resulting histogram would be $\langle 3, 0, 9 \rangle$, since there are 3 cells in empty columns of height 1, no empty cells in columns of height 2, and 9 cells in empty columns of height 3.

Assume now that we only have left to place a $2 \times 3$ and a $2 \times 2$ rectangle. We can assign the six cells of the $2 \times 3$ rectangle to the empty cells of $v_3{=}9$, leaving us with the remaining empty cells $\langle 3, 0, 3 \rangle$. At this point, we cannot assign the area of the $2 \times 2$, because we only have 3 empty cells that can accommodate its height and we need 4, so we can prune.

In general, for a set of unplaced rectangles $R$ and a bounding box of height $H$,

$$\forall h, \left[ \sum_{r \in R, h_r \geq h} w_r h_r \leq \sum_{i=h}^{H} v_i \right], \tag{1}$$

where a rectangle $r \in R$ has dimensions $w_r \times h_r$. That is, for every given height $h$, the amount of space that can accommodate rectangles of height $h$ or greater must be at least the cumulative area of rectangles of height $h$ or greater. We check this constraint after each $x$-coordinate assignment.

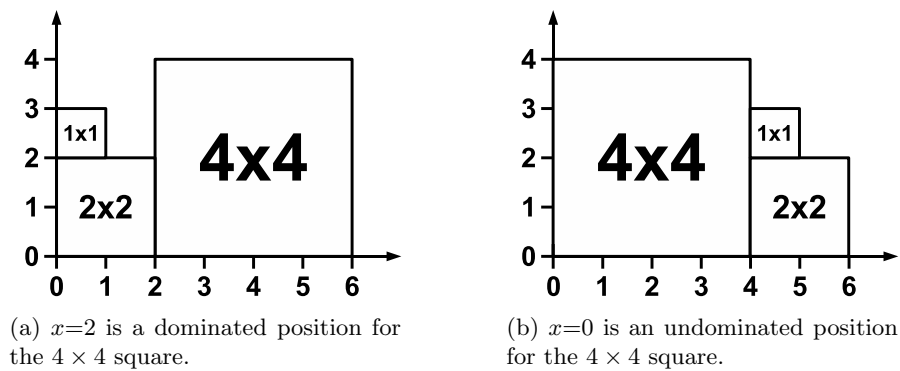

(a) $x$=2 is a dominated position for the $4 \times 4$ square.

(b) $x$=0 is an undominated position for the $4 \times 4$ square.

Figure 5: Example of dominance conditions.

### 3.5.2 Pruning With Dominance Conditions

Korf (2003) introduced a set of dominance conditions to prune positions where large rectangles are too close to the sides of the bounding box. For example, imagine that we must pack the squares $4 \times 4$, $3 \times 3$, $2 \times 2$, and $1 \times 1$. In Figure 5a, the placement of the $4 \times 4$ square leaves a $2 \times 4$ gap against the left side of the bounding box in which the $3 \times 3$ square cannot fit. Only the $2 \times 2$ and $1 \times 1$ squares can fit within the gap, and in fact they both can be placed entirely within the gap. Notice that in any solution with an arrangement as in Figure 5a, we can always rearrange them as in Figure 5b without disturbing any other squares. Thus, there is no need to try placing the $4 \times 4$ square at $x$=2 so long as we have tried placing it at $x$=0. In general, a rectangle placement is dominated if it leaves a gap in which all rectangles that can individually fit can also be packed together in the gap without protruding from it. Although Korf hard-coded dominance rules for the consecutive-square benchmark, we dynamically generate them for every instance with insignificant preprocessing overhead.

### 3.5.3 Variable Ordering

In the following subsections we consider two variable orders that work together in our packer. We use a fixed ordering that governs which rectangle is assigned next. This ordering is used for the $x$-intervals independently from its use on the single $x$-coordinate variables. At any point in time, we also must choose whether to assign the next $x$-interval or the next single $x$-coordinate variable. Since the ordering between $x$-intervals and single $x$-coordinate variables is simpler, we present this technique first.

**Ordering Between X-Intervals and X-Coordinates By Area**  Our variable order is based on the observation that placing rectangles of larger area is more constraining than placing those of smaller area. At all times we can either choose to assign a single $x$-coordinate to a rectangle for which we previously had assigned an $x$-interval, or we can assign an $x$-interval to a rectangle we have not yet made any assignments for. As shown in Figure 4, either of these assignments will decrease the amount of empty space represented in the cumulative constraint vector. We always pick next the variable that results in the least remaining space.

59

**Ordering Among Rectangles By Branching Factor** There is a natural variable order that arises from both the consecutive-square and unoriented consecutive-rectangle benchmarks when using the strategy of picking the most constrained variable next. For example, in the consecutive-square benchmark, the largest rectangle is clearly the largest in height, width, and area. However, in our new benchmarks the rectangle of largest width has the smallest height, but not the largest area, making a good variable ordering non-obvious.

We propose a variable order over rectangles of various aspect ratios by picking the variable with the fewest number of values first, to favor a smaller branching factor closer to the root of the search tree. For the oriented equal-perimeter benchmark, recall that we assign intervals to the $x$-coordinates before the individual $x$-coordinates, and like Simonis and Sullivan (2008) we use a constant factor times the rectangle width to define the interval size. The branching factor for the $x$-interval variables for a given rectangle is

$$b = \frac{B_w - r_w}{Cr_w} = \frac{B_w}{C}\left[\frac{1}{r_w}\right] - \frac{1}{C}, \tag{2}$$

where $B_w$ is the bounding box width, $r_w$ is the rectangle width, and $C$ is a constant chosen experimentally. The numerator $B_w - r_w$ is the number of $x$-coordinate values that the rectangle can have while still fitting in the bounding box, and the denominator $Cr_w$ is the size of the interval we will be assigning to the given rectangle. For example, if $C$=0.75 then we would assign intervals of size three to a $4 \times 2$ rectangle.

We may drop the translational constant $-1/C$ as well as the positive scalar $B_w/C$ since we are only interested in a relative ordering for the rectangles, leaving us with $1/r_w$ which means that for the oriented benchmark we should place the rectangles in order of decreasing width. For the unoriented double-perimeter benchmark, our packer first tries all values for a particular $x$-interval, and then rotates the rectangle 90-degrees before trying another set of $x$-interval values. In this case the branching factor is

$$b = \frac{B_w - r_w}{Cr_w} + \frac{B_w - r_h}{Cr_h} = \frac{B_w}{C}\left[\frac{1}{r_w} + \frac{1}{r_h}\right] - \frac{2}{C}. \tag{3}$$

As mentioned before, we can drop the scalar and translational constant, giving us

$$\frac{1}{r_w} + \frac{1}{r_h} = \frac{r_w + r_h}{r_w r_h}. \tag{4}$$

Because all rectangles in a given instance have the same perimeter by definition, the numerator of the result in Equation 4 is constant. Therefore for our unoriented benchmark, we place the rectangles in order of decreasing area.

### 3.5.4 Determining Sizes of $X$-Intervals

On the consecutive-square benchmark, our packer used an interval size that is 0.35 times the width of a given rectangle. We found that larger interval sizes improve the performance of our packer on the new equal-perimeter benchmarks, and use a value of $C$=0.55 instead.

As we assign larger intervals to the short and wide rectangles, the $x$-interval variables for these rectangles tend to have branching factors of three or less. We should balance the sizes of these intervals so that the values assigned are equally constraining on their subtrees. For example, consider $C$=0.55, a rectangle of width 20, and its set of possible $x$-coordinate values [0,23]. Without balancing the sizes of the intervals, our packer would explore interval sizes of $20C = 11$, such as $x$=[0,10], $x$=[11,21], and finally the remaining domain values with a small interval of $x$=[22,23]. This results in small compulsory parts and therefore large search subtrees in the first two branches, but a very large compulsory part and thus a small search subtree in the third.

Since we must explore three branches anyway, we can balance the sizes of these interval assignments by exploring $x$=[0,7], $x$=[8,15], and $x$=[16,23]. The eventual effect is a better balance on the size of the search subtrees amongst branches. Our packer first computes the branching factor induced by the global interval parameter $C$=0.55 for each rectangle, and then it balances the number of values in each interval assignment.

**Interactions Between Interval Assignment and Dominance Conditions**   On consecutive-square instances, for most of the squares there are several positions following $x$=0 that are dominated. Therefore, our packer first branches by assigning the degenerate interval $x$=[0,0] before exploring interval assignments for the undominated positions. Although this technique increased the performance of our packer fivefold compared to leaving it out, the same strategy slowed the performance fivefold on the oriented and unoriented double-perimeter benchmark. The reason for this degradation of performance is as follows.

In our equal-perimeter benchmarks, the $1 \times N$ rectangle can always partially fit in gaps left by other rectangles, but it must always protrude out of those gaps, thereby eliminating the dominance conditions we previously described. Without any dominated positions to account for, simply applying the same strategy used for consecutive-squares on our new benchmarks results in our packer committing to single $x$-coordinate values in situations where it is more desirable to include those positions in a larger interval assignment. To avoid this, our packer detects when there are no dominated positions and dynamically chooses whether to assign the degenerate interval as the $x$-coordinate assignment, or to immediately begin with interval assignments.

## 3.6 Perfect Packing Transformation

For every complete $x$-coordinate solution, we transform the problem instance into a perfect packing problem instance before working on the $y$-coordinates. A perfect packing instance is a rectangle packing problem with the property that the solution has no empty space. The transformation is done by adding to the original set of rectangles a number of $1 \times 1$ rectangles necessary to increase the total area of the rectangles to that of the bounding box. Although the new $1 \times 1$ rectangles increase the problem size, the hope is that the ease of solving perfect packing instances will offset the difficulty of packing more rectangles. Next we describe our search space for perfect packing. As we will show, our methods rely on the perfect packing property of having no empty space.

## 3.7 Assigning Y-Coordinates

An alternative to asking "Where should this rectangle go?" is to ask "Which rectangle should go here?" In the former model, rectangles are variables and empty locations are values, whereas in the latter, empty locations are variables and rectangles are values. For $y$-coordinates, we search the latter model. We use a 2D bitmap to draw in placed rectangles to test for overlap, and we backtrack on positions that cannot accommodate any remaining rectangles, or as required by Korf's (2003) wasted space pruning rule.

### 3.7.1 Empty Corner Model

In all perfect packing solutions, every rectangle's lower-left corner fits in some lower-left empty corner formed by other rectangles, the sides of the bounding box, or a combination of both. In this model, we have one variable per empty corner. In the final solution, since each rectangle goes into exactly one empty corner, the number of empty corner variables is equal to the number of rectangles in the perfect packing instance. The set of values is just the set of unplaced rectangles.

This search space has the interesting property that variables are dynamically created during search because the $x$- and $y$-coordinates of an empty corner are known only after the rectangles that create it are placed. Furthermore, placing a rectangle in an empty corner assigns both its $x$- and $y$-coordinates.

Note that the empty corner model can describe all perfect packing solutions. Given any perfect packing solution, we can list a unique sequence of all the rectangles by scanning left to right, bottom to top for the lower-left corners of the rectangles. This sequence corresponds to a sequence of assignments from the root of this search space to a leaf in the tree. This property also bounds the maximum size of the search space by $N'!$ where $N'$ is the number of rectangles after we have performed the perfect packing transformation.

### 3.7.2 Duplicate Rectangles

Due to the additional $1 \times 1$ rectangles from the perfect packing transformation, we have introduced additional redundancy into the problem. A simple way to handle this is as follows. For a particular empty corner, we never place a rectangle that is a duplicate of one we have already tried at that position. This method of handling duplicates also applies to duplicate rectangles in the original problem instance.

## 4. Experimental Results

We benchmarked our packers in Linux on a 2GHz AMD Opteron 246 with 2GB of RAM. The packer we call KMP10 (Korf et al., 2010) was benchmarked on the same machine, so we quote their published results. We do not include data for their relative placement packer because it was not competitive. Results for Simonis and O'Sullivan's packer (2008), which we call SS08, are also quoted, obtained from SICStus Prolog 4.0.2 for Windows on a 3GHz Intel Xeon 5450 with 3.25GB of RAM. Since their machine is faster than ours, these comparisons are a conservative estimate of our relative performance.

### 4.1 Previous Benchmarks

Because both the consecutive-square benchmark and the unoriented consecutive-rectangle benchmarks (Korf et al., 2010) have been used in the literature to measure performance, we include data collected using these two benchmarks.

| Size $N$ | KMP10 Time | SS08 Time | FixedOrder Time | HK09 Time |
|---|---|---|---|---|
| 20 | 1:32 | :02 | :00 | :00 |
| 21 | 9:54 | :07 | :03 | :03 |
| 22 | 37:03 | :51 | :02 | :02 |
| 23 | 3:15:23 | 3:58 | :14 | :12 |
| 24 | 10:17:02 | 5:56 | :40 | :37 |
| 25 | 2:02:58:36 | 40:38 | 2:27 | 2:14 |
| 26 | 8:20:14:51 | 3:41:43 | 10:25 | 9:39 |
| 27 | 34:04:01:03 | 11:30:02 | 1:08:55 | 35:12 |
| 28 | | | 2:18:12:13 | 4:39:31 |
| 29 | | | | 8:06:03 |
| 30 | | | | 2:17:32:52 |
| 31 | | | | 4:16:03:42 |
| 32 | | | | 33:11:36:23 |

Table 1: CPU times required by various packers on the consecutive-square benchmark, where the task is to pack squares from $1 \times 1$ up to $N \times N$.

### 4.1.1 Consecutive Squares

Table 1 compares the CPU runtimes of four packers on the consecutive-square benchmark. The first column specifies the instance size, which is both the number of squares and the size of the largest one. The remaining columns specify the CPU times required by various algorithms to find all the optimal solutions in the format of days, hours, minutes, and seconds. When there are multiple boxes of minimum area, as for $N$=27 as listed in Table 8 of Appendix 4.4, we report the total time required to find all optimal bounding boxes. We do this for two reasons. First, finding all minimum area bounding boxes removes the question of which bounding box to test first if more than one have the same area. Second, by providing all optimal solutions, other researchers working on rectangle packing can use this information to verify the correctness of their programs.

HK09 includes our wasted space pruning rule for the $x$-coordinates, dynamic variable ordering between $x$-intervals and $x$-coordinates, the perfect packing transformation, and its related search space and inference rules. We have named this packer to be consistent with our previous work (Huang & Korf, 2009). SS08 refers to the previous state-of-the-art packer (Simonis & O'Sullivan, 2008). The largest problem previously solved was $N$=27 and took SS08 over 11 hours. We solved the same problem in 35 minutes and solved five more open problems up to $N$=32. KMP10 refers to Korf et al.'s (2010) absolute placement packer. FixedOrder assigns all $x$-intervals before any single $x$-coordinates, but includes all of our other ideas. HK09's dynamic variable ordering for the $x$-coordinates was an order of magnitude faster than FixedOrder by $N$=28. The order of magnitude improvement of FixedOrder over SS08 is likely due to our use of perfect packing for assigning the $y$-coordinates. We do not include the timing for a packer with perfect packing disabled because it was not competitive (e.g., $N$=20 took over 2.5 hours).

| Size $N$ | X-Coordinate Solutions | Seconds in X | Seconds in Y | Ratio X:Y |
|---|---|---|---|---|
| 21 | 665 | 0.35 | 1.04 | 0.3 |
| 22 | 283 | 0.95 | 0.18 | 5.3 |
| 23 | 391 | 6.54 | 0.31 | 21.1 |
| 24 | 870 | 19.41 | 1.08 | 18.0 |
| 25 | 193 | 73.38 | 0.14 | 524.1 |
| 26 | 1,026 | 313.81 | 1.39 | 225.8 |
| 27 | 244 | 1,181.53 | 0.60 | 1,969.2 |
| 28 | 2,715 | 8,987.36 | 23.40 | 384.1 |
| 29 | 11,129 | 15,677.20 | 28.82 | 544.0 |
| 30 | 10,244 | 124,399.74 | 17.97 | 6,922.6 |
| 31 | 73,614 | 214,575.08 | 254.42 | 843.4 |
| 32 | 37,742 | 1,916,312.67 | 102.59 | 18,679.3 |

Table 2: CPU times spent searching for $x$- and $y$-coordinates for the consecutive-square benchmark

In Table 2 the second column is the number of complete $x$-coordinate assignments our packer found over the entire run of a particular problem instance. The third column is the total time spent in searching for the $x$-coordinates. The fourth column is the total time spent in performing the perfect packing transformation and searching for the $y$-coordinates. Both columns represent the total CPU time over an entire run for a given problem instance. The last column is the ratio of time in the third column to that of the fourth. Interestingly, almost all of the time is spent on the $x$-coordinates as opposed to the $y$-coordinates, which suggests that if we could efficiently enumerate the $x$-coordinate solutions, we could also efficiently solve rectangle packing. This is confirmed by the relatively few $x$-coordinate solutions that exist even for large instances. The data in Table 2 was obtained on a Linux 2.93GHz Intel Core 2 Duo E7500 machine in a separate experiment from that of Table 1, which is why the total time spent on a given instance is different.

### 4.1.2 Unoriented Consecutive Rectangles

Table 3 compares the CPU times of our packer on the unoriented consecutive-rectangles benchmark with that of Korf et al. (2010). Although the techniques due to Simonis and O'Sullivan (2008) outperform those of Korf et al. on the consecutive-square benchmark, there were no previously published results on this benchmark besides that of Korf et al. Because this benchmark is easier than the consecutive-square benchmark, we do not break down the contributions of each of our techniques, as these differences were delineated more clearly in the previous section. The primary differentiating feature of this benchmark is that rectangles are unoriented.

The first column gives the size of the problem instance. The second column gives the performance of the previous state-of-the-art packer on this benchmark, using Korf et al.'s code (2010). The third column gives the performance of our packer on this benchmark. All of the data in this table was collected on a Linux 2.93GHz Intel Core 2 Duo E7500 machine, except for $N$=28 and $N$=29, which were collected on a Linux 2.53GHz Intel Xeon E5630 with 12GB of RAM, and which our experiments revealed to be 20% faster than the former machine.

| Size $N$ | KMP10 Time | HK10 Time |
|---|---|---|
| 16 | :01 | :00 |
| 17 | :05 | :00 |
| 18 | :17 | :00 |
| 19 | :07 | :00 |
| 20 | 8:11 | :05 |
| 21 | 15:00 | :06 |
| 22 | 1:09:45 | :17 |
| 23 | 8:51:46 | :47 |
| 24 | 11:53:17 | 13:38 |
| 25 | 7:17:00:03 | 2:21:10 |
| 26 | | 6:31:51 |
| 27 | | 4:07:37:08 |
| 28 | | 1:16:43:02 |
| 29 | | 6:04:47:06 |

Table 3: CPU times required by two packers on the unoriented consecutive-rectangle benchmark, where the task is to pack unoriented rectangles of sizes $1{\times}2$, $2{\times}3$, ..., and $N{\times}(N{+}1)$.

For this benchmark our techniques have allowed us to extend the known solutions from $N$=25 to $N$=29 and allowed us to solve $N$=25 about 80 times faster than the previous state-of-the-art on this benchmark.

## 4.2 Oriented Equal-Perimeter and Unoriented Double-Perimeter Rectangles

This section uses our new benchmarks to compare the techniques we have developed for non-square instances. The techniques we discuss here, including the dynamic adjustment of interval sizes and the generalized variable order based on branching factor, largely do not affect the performance of our packer on the consecutive-square benchmark. In fact, we tested this packer on that benchmark to see the effects of any extra overhead added by our improvements. Our new packer resulted in only a five percent speedup compared to our packer without these changes on the consecutive-square benchmark, likely due to minor improvements in data structures, and balancing interval sizes. Therefore, we compare the effects of these techniques only on our new benchmarks. Because the techniques we have developed for our new benchmarks improve performance in both the oriented and unoriented cases, we discuss them together.

Table 4 compares the performance of our packers on the oriented equal-perimeter benchmark while Table 5 compares the same packers using our unoriented double-perimeter benchmark. The first column refers to the problem size of the instance, which is the number of rectangles. The second column gives the number of bounding boxes tested in order to find all optimal solutions. The remaining columns represent the CPU times for different versions of our packer in the format of days, hours, minutes, and seconds. We wrote our packer in C++ and collected our data on a Linux 2.93GHz Intel Core 2 Duo E7500 machine.

| Size $N$ | Boxes Tested | HK09 Time | OptDom Time | BrFactor Time | C=0.55 Time | HK10 Time |
|---|---|---|---|---|---|---|
| 13 | 7 | :01 | :00 | :00 | :00 | :00 |
| 14 | 7 | :02 | :01 | :00 | :00 | :00 |
| 15 | 10 | :16 | :05 | :01 | :00 | :00 |
| 16 | 9 | :57 | :16 | :02 | :00 | :00 |
| 17 | 8 | 5:56 | 1:21 | :27 | :03 | :02 |
| 18 | 12 | 1:06:32 | 14:47 | 6:15 | :32 | :22 |
| 19 | 12 | 6:35:48 | 1:26:16 | 31:23 | 3:34 | 2:15 |
| 20 | 11 | 1:18:51:34 | 7:36:09 | 1:51:10 | 13:06 | 7:51 |
| 21 | 9 | 3:21:31:46 | 13:33:16 | 4:22:49 | 20:49 | 11:20 |
| 22 | 15 | | | | 14:22:03 | 9:12:37 |
| 23 | 16 | | | | | 3:22:50:38 |

Table 4: CPU times required by various packers on the oriented equal-perimeter rectangle benchmark, where the task is to pack oriented rectangles of sizes $1 \times N$, $2 \times (N-1)$, ..., $(N-1) \times 2$, and $N \times 1$.

| Size $N$ | Boxes Tested | HK09 Time | OptDom Time | BrFactor Time | C=0.55 Time | HK10 Time |
|---|---|---|---|---|---|---|
| 11 | 12 | :01 | :00 | :00 | :00 | :00 |
| 12 | 17 | :20 | :04 | :04 | :01 | :01 |
| 13 | 13 | 1:45 | :21 | :21 | :06 | :06 |
| 14 | 17 | 28:48 | 4:53 | 4:53 | 1:19 | 1:15 |
| 15 | 21 | 1:43:01 | 11:36 | 11:36 | 3:33 | 2:34 |
| 16 | 35 | 1:16:46:44 | 4:13:34 | 4:13:34 | 1:16:02 | 1:01:54 |
| 17 | 27 | | 1:12:40:14 | 1:12:40:14 | 9:44:14 | 7:53:50 |
| 18 | 35 | | | | | 2:02:10:38 |

Table 5: CPU times required by various packers on the unoriented double-perimeter rectangle benchmark, where the task is to pack unoriented rectangles of sizes $1 \times (2N-1)$, $2 \times (2N-2)$, ..., $(N-1) \times (N+1)$, and $N \times N$.

From left to right, each successive packer improves on the previous one by including an additional technique. The column called HK09 is data collected using only the techniques developed specifically for consecutive-square packing, which include the perfect packing transformation and its related inference rules, dynamic variable ordering between single $x$-coordinates and $x$-intervals, and the wasted space pruning rule for the $x$-coordinates (Huang & Korf, 2009). To compare against our new variable ordering over rectangles of various aspect ratios, we used the order of decreasing area by default in HK09.

OptDom improves upon HK09 by dynamically detecting when dominance rules apply or are inapplicable, and optimizes the $x$-interval assignments with this knowledge. BrFactor improves upon OptDom in that it orders the oriented equal-perimeter benchmark by decreasing width and the unoriented double-perimeter benchmark by decreasing area. $C$=0.55 improves upon BrFactor in that we use an interval size of 0.55 instead of $C$=0.35 as we did for the consecutive-square benchmark. Finally, HK10 improves upon $C$=0.55 by using knowledge of the branching factor to rebalance the sizes of the interval assignments for the $x$-coordinates.

Notice that OptDom, BrFactor, and $C$=0.55 introduce techniques that reduce the branching factor, and so they have a greater effect on performance than HK10, whose new technique seeks to make the intervals assigned equally constraining. Our experiments reveal that these techniques interact with one another, and we note that without including dominated positions in the intervals, the performance gained from the other techniques appears muted. This interaction is also why we tune the global interval parameter $C$ only after including the other techniques that affect the branching factor.

Ordering by branching factor improved performance for our oriented equal-perimeter benchmark but not for our unoriented benchmark. In the latter case, as seen in Table 5, our technique of ordering by branching factor prescribes ordering by decreasing area, which is what we gave the packer as a reasonable default. Therefore, there is no difference in the algorithm nor in its performance between the OptDom and BrFactor columns of Table 5.

Note that the unoriented double-perimeter benchmark requires our packer to try over twice as many bounding boxes for a given parameter $N$ than that required for our oriented benchmark. This is due to having $2N$-1 as the largest dimension in the unoriented benchmark while having $N$ as the largest dimension in the oriented benchmark. The larger rectangles introduce a higher precision into the problem, and so we must try more bounding boxes. The containment problem for an unoriented instance has a problem space that is a factor of $2^N$ larger than that of an oriented instance due to the two orientations of each rectangle. Thus, an instance with $N$ rectangles in this benchmark is incomparable to an instance of $N$ squares from the consecutive-square benchmark when evaluating benchmark difficulty.

In summary, using all of our techniques together, we can solve $N$=21 of the oriented equal-perimeter benchmark about 500 times faster and $N$=16 of the unoriented double-perimeter benchmark about 40 times faster than the techniques we presented optimized only for consecutive squares.

| Size | Boxes Tested | | CPU Time | |
|---|---|---|---|---|
| $N$ | Squares | Perimeter | Squares | Perimeter |
| 16 | 10 | 9 | :00 | :00 |
| 17 | 5 | 8 | :00 | :02 |
| 18 | 14 | 12 | :00 | :22 |
| 19 | 12 | 12 | :00 | 2:15 |
| 20 | 14 | 11 | :00 | 7:51 |
| 21 | 20 | 9 | :01 | 11:20 |
| 22 | 17 | 15 | :01 | 9:12:37 |
| 23 | 19 | 16 | :07 | 3:22:50:38 |
| 24 | 19 | | :20 | |
| 25 | 17 | | 1:14 | |
| 26 | 21 | | 5:15 | |
| 27 | 22 | | 19:42 | |
| 28 | 30 | | 2:30:11 | |
| 29 | 27 | | 4:21:46 | |
| 30 | 21 | | 1:10:33:38 | |
| 31 | 30 | | 2:11:40:29 | |
| 32 | 36 | | 22:04:20:15 | |

Table 6: Number of bounding boxes tested and CPU time required to solve a given instance in the consecutive-square and the oriented equal-perimeter benchmarks.

## 4.3 Comparing Easy and Hard Benchmarks

The following tables compare the difficulty of various benchmarks using our packer (Huang & Korf, 2010) with all optimizations enabled.

### 4.3.1 Consecutive Squares vs. Equal-Perimeter Rectangles

In Table 6, the first column indicates the number of rectangles in the instance. The second and third columns labeled "Boxes Tested" give the number of bounding boxes that were tested when finding all optimal solutions for the consecutive-square benchmark and the oriented equal-perimeter benchmark, respectively. The fourth and fifth columns give the performance of our rectangle packer on both benchmarks as well. Each data point in this table was collected using a Linux 2.93GHz Intel Core 2 Duo E7500 using one process, one thread, and one core.

Notice that for a given instance with the same number of rectangles, the oriented equal-perimeter benchmark is significantly harder than the consecutive-square benchmark. This is due to the fact that for a given problem size, the consecutive-square benchmark contains many little squares that are typically easy to place – a property missing in the equal-perimeter benchmark. In fact, by $N=23$ our packer requires over four orders of magnitude more time to find the optimal solutions to our new benchmark compared to an instance with the same number of items from the consecutive-square benchmark.

### 4.3.2 Unoriented Consecutive-Rectangles vs. Unoriented Double-Perimeter Rectangles

Table 7 shows how removing certain properties results in successively more difficult benchmarks. We start with the unoriented consecutive-rectangle benchmark (Korf et al., 2010) which contains many easy properties. In the "Doubly Scaled" column we pack $2 \times 4$, $4 \times 6$, $6 \times 8$, ..., $(2N) \times (2N+2)$ rectangles, which simply scales up the unoriented consecutive-rectangle benchmark by a factor of two. This benchmark is more difficult because integers of higher magnitude lead to more $x$-coordinates to search, which in turn increases the branching factor of the problem. In the "Unique Dimensions" column we now pack rectangles of sizes $1 \times 2$, $3 \times 4$, $5 \times 6$, ..., $(2N-1) \times (2N)$, which differs from the previous benchmark in that all dimensions are unique. The last column distributes the area among the rectangles more uniformly to avoid consolidating most of the area in the first few rectangles. This column is also the culmination of all of the difficult properties which we have identified for a rectangle packing benchmark, which we call our unoriented double-perimeter benchmark. All data points in this table were collected using a Linux 2.93GHz Intel Core 2 Duo E7500 machine without any parallelization, except for $N=28$ and $N=29$, which were collected on a Linux 2.53GHz Intel Xeon E5630 machine with 12GB of RAM, which we estimate to be thirty percent faster.

## 4.4 Bounding Boxes of Minimum Area

In this section we list all of the optimal bounding boxes on various benchmarks found by our program with all optimizations enabled. Notice that we do not duplicate the data for the unoriented high-precision rectangle benchmark, and leave it in Table 10, Section 5.5.2, since the discussion there refers to this data.

| Size $N$ | Unoriented Consecutive-Rectangles | Doubly Scaled | Unique Dimensions | Unoriented Double-Perimeter |
|---|---|---|---|---|
| 12 | :00 | :00 | :00 | :01 |
| 13 | :00 | :00 | :00 | :06 |
| 14 | :00 | :00 | :01 | 1:15 |
| 15 | :00 | :00 | :00 | 2:34 |
| 16 | :00 | :00 | :01 | 1:01:54 |
| 17 | :00 | :00 | :01 | 7:53:50 |
| 18 | :00 | :01 | :03 | 2:02:10:38 |
| 19 | :00 | :01 | :11 | |
| 20 | :05 | :09 | :50 | |
| 21 | :06 | :10 | 3:00 | |
| 22 | :17 | :29 | 15:34 | |
| 23 | :47 | 1:13 | 3:21:36 | |
| 24 | 13:38 | 27:37 | 12:23:37 | |
| 25 | 2:21:10 | 6:41:20 | | |
| 26 | 6:31:51 | 1:02:12:06 | | |
| 27 | 4:07:37:08 | | | |
| 28 | 1:16:43:02 | | | |
| 29 | 6:04:47:06 | | | |

Table 7: CPU time required for our optimized packer on various benchmarks of increasing difficulty.

The first column in tables 8 and 9 refer to the size of the problem instance for their respective benchmarks. The columns called Optimal Solution give the dimensions of the optimal bounding boxes for a given instance. The next column called Empty Space gives the percent of empty space in the optimal solution. The next column gives the number of bounding boxes that were tested in order to find all optimal solutions for a given instance.

## 5. Absolute Placement on High-Precision Instances

Meir and Moser (1968) first proposed the problem of finding the smallest square that can contain an infinite series of rectangles of sizes $\frac{1}{1} \times \frac{1}{2}, \frac{1}{2} \times \frac{1}{3}, \frac{1}{3} \times \frac{1}{4}$, ..., etc. The rectangles cannot overlap and are unoriented. The unit square has exactly enough area since the total area of the rectangles in the infinite series is one. On the other hand, no space can be wasted, suggesting that the task is infeasible. Inspired by this problem, we propose our last benchmark and developed several new techniques.

We introduce the *unoriented high-precision rectangle benchmark*, where the task is to find all bounding boxes of minimum area that can contain a finite set of unoriented rectangles of sizes $\frac{1}{1} \times \frac{1}{2}, \frac{1}{2} \times \frac{1}{3}$, ..., up to $\frac{1}{N} \times \frac{1}{N+1}$. For example, for $N$=4 one must pack rectangles of sizes $\frac{1}{1} \times \frac{1}{2}, \frac{1}{2} \times \frac{1}{3}, \frac{1}{3} \times \frac{1}{4}$, and $\frac{1}{4} \times \frac{1}{5}$. Alternatively, one may try to pack rectangles of sizes $60 \times 30$, $30 \times 20$, $20 \times 15$, and $15 \times 12$ into a $60 \times 60$ square, which is just the original instance scaled up by a factor of 60, the least common multiple of the rectangle denominators. This strategy is required for the broad class of recent rectangle-packers that explore the domain of integer $x$- and $y$-coordinates for the rectangles and quickly break down at higher $N$. For example, the optimal solution of $N$=15 has over 400 billion unique coordinate pairs that rectangles can be assigned to. Our benchmark complements rather than replaces the current low-precision benchmarks, which until now have neglected high-precision instances.

The remainder of this section is organized as follows. We first review some of the previous work proposing solution techniques that may be unaffected by the precision of the rectangle dimensions. Then we describe several adaptations of our low-precision techniques to the high-precision case, along with some new techniques developed specifically for high-precision rectangle instances, and finally follow with experimental results.

### 5.1 Previous Work

The relative placement approach of Moffitt and Pollack (2006) for rectangle packing, and similar types of search spaces used in resource-constrained scheduling (Weglarz, 1999), promises to be immune to the problem of high-precision rectangle instances. However, since there are so many techniques that we have described in the previous sections that cannot be extended to a packer working in the relative placement search space, we have decided to stay within the absolute placement framework and attempt to mitigate the problems introduced by high-precision numbers.

| | Consecutive Squares | | | Consecutive Rectangles | | |
|---|---|---|---|---|---|---|
| Size $N$ | Optimal Solutions | Empty Space | Boxes Tested | Optimal Solutions | Empty Space | Boxes Tested |
| 1 | 1×1 | 0.00% | 1 | 1×2 | 0.00% | 1 |
| 2 | 2×3 | 16.7% | 1 | 2×4 | 0.00% | 1 |
| 3 | 3×5 | 6.67% | 1 | 4×5 | 0.00% | 1 |
| 4 | 5×7 | 14.3% | 1 | 5×8, 4×10 | 0.00% | 2 |
| 5 | 5×12 | 8.33% | 1 | 5×14 | 0.00% | 2 |
| 6 | 9×11 | 8.08% | 1 | 6×19 | 1.75% | 2 |
| 7 | 11×14, 7×22 | 9.09% | 3 | 12×14 | 0.00% | 2 |
| 8 | 14×15 | 2.86% | 2 | 15×16 | 0.00% | 1 |
| 9 | 15×20 | 5.00% | 4 | 16×21, 14×24 | 1.79% | 5 |
| 10 | 15×27 | 4.94% | 5 | 17×26 | 0.45% | 5 |
| 11 | 19×27 | 1.36% | 3 | 22×26 | 0.00% | 2 |
| 12 | 23×29 | 2.55% | 6 | 21×35 | 0.95% | 4 |
| 13 | 22×38 | 2.03% | 5 | 26×35 | 0.00% | 1 |
| 14 | 23×45 | 1.93% | 8 | 32×35, 28×40 | 0.00% | 2 |
| 15 | 23×55 | 1.98% | 13 | 34×40 | 0.00% | 1 |
| 16 | 28×54, 27×56 | 1.06% | 10 | 32×51 | 0.00% | 2 |
| 17 | 39×46 | 0.50% | 5 | 34×57 | 0.00% | 2 |
| 18 | 31×69 | 1.40% | 14 | 30×76 | 0.00% | 3 |
| 19 | 47×53 | 0.84% | 12 | 35×76, 38×70 | 0.00% | 2 |
| 20 | 34×85 | 0.69% | 14 | 35×88, 44×70, 55×56 | 0.00% | 4 |
| 21 | 38×88 | 0.99% | 20 | 39×91 | 0.20% | 2 |
| 22 | 39×98 | 0.71% | 17 | 44×92 | 0.00% | 2 |
| 23 | 64×68 | 0.64% | 19 | 40×115, 46×100 | 0.00% | 3 |
| 24 | 56×88 | 0.58% | 19 | 40×130, 52×100, 65×80 | 0.00% | 4 |
| 25 | 43×129 | 0.40% | 17 | 45×130, 65×90, 75×78 | 0.00% | 5 |
| 26 | 70×89 | 0.47% | 21 | 42×156, 52×126, 56×117, 63×104, 72×91, 78×84 | 0.00% | 7 |
| 27 | 47×148, 74×94 | 0.37% | 22 | 63×116 | 0.00% | 3 |
| 28 | 63×123 | 0.45% | 30 | 56×145, 70×116 | 0.00% | 3 |
| 29 | 81×106 | 0.36% | 27 | 62×145 | 0.00% | 2 |
| 30 | 51×186 | 0.33% | 21 | | | |
| 31 | 91×110 | 0.33% | 30 | | | |
| 32 | 85×135 | 0.31% | 36 | | | |

Table 8: The optimal solutions for the consecutive-square benchmark, where the task is to pack squares of sizes 1×1, 2×2, ..., and $N×N$, and for the unoriented consecutive-rectangle benchmark, where the task is to pack unoriented rectangles of sizes $1 × 2$, $2 × 3$, ..., and $N × (N+1)$.

| Size $N$ | Oriented Equal Perimeter | | | Unoriented Double Perimeter | | |
|---|---|---|---|---|---|---|
| | Optimal Solutions | Empty Space | Boxes Tested | Optimal Solutions | Empty Space | Boxes Tested |
| 1 | 1×1 | 0.00% | 1 | 1×1 | 0.00% | 1 |
| 2 | 2×3 | 33.3% | 1 | 3×3 | 22.2% | 1 |
| 3 | 3×4 | 16.7% | 1 | 3×8 | 8.33% | 2 |
| 4 | 4×6 | 16.7% | 1 | 6×9 | 7.41% | 2 |
| 5 | 6×7 | 16.7% | 4 | 6×17 | 6.86% | 8 |
| 6 | 6×10 | 6.67% | 2 | 9×19 | 5.85% | 9 |
| 7 | 8×11 | 4.55% | 2 | 13×20 | 3.08% | 11 |
| 8 | 8×16 | 6.25% | 5 | 18×21 | 1.59% | 8 |
| 9 | 11×16 | 6.25% | 6 | 13×41 | 1.50% | 13 |
| 10 | 11×21 | 4.76% | 8 | 24×30 | 0.69% | 8 |
| 11 | 14×21 | 2.72% | 6 | 29×33 | 1.15% | 12 |
| 12 | 13×29 | 3.45% | 7 | 21×59 | 1.37% | 17 |
| 13 | 16×29 | 1.94% | 7 | 38×41 | 0.71% | 13 |
| 14 | 19×30, 15×38 | 1.75% | 7 | 38×51, 17×114 | 0.67% | 17 |
| 15 | 24×29 | 2.30% | 10 | 44×54 | 0.67% | 21 |
| 16 | 23×36 | 1.45% | 9 | 45×64, 30×96, 40×72, 48×60 | 0.83% | 35 |
| 17 | 24×41 | 1.52% | 8 | 39×88, 52×66 | 0.44% | 27 |
| 18 | 24×48 | 1.04% | 12 | 55×74 | 0.57% | 35 |
| 19 | 32×42, 24×56 | 1.04% | 12 | | | |
| 20 | 37×42 | 0.90% | 11 | | | |
| 21 | 35×51 | 0.78% | 9 | | | |
| 22 | 34×60 | 0.78% | 15 | | | |
| 23 | 38×61 | 0.78% | 16 | | | |

Table 9: The optimal solutions to the oriented equal-perimeter rectangle benchmark, where the task is to pack oriented rectangles of sizes $1 \times N$, $2 \times (N-1)$, ..., $(N-1) \times 2$, and $N \times 1$, and to the unoriented double-perimeter rectangle benchmark, where the task is to pack unoriented rectangles of sizes $1 \times (2N-1)$, $2 \times (2N-2)$, ..., $(N-1) \times (N+1)$, and $N \times N$.

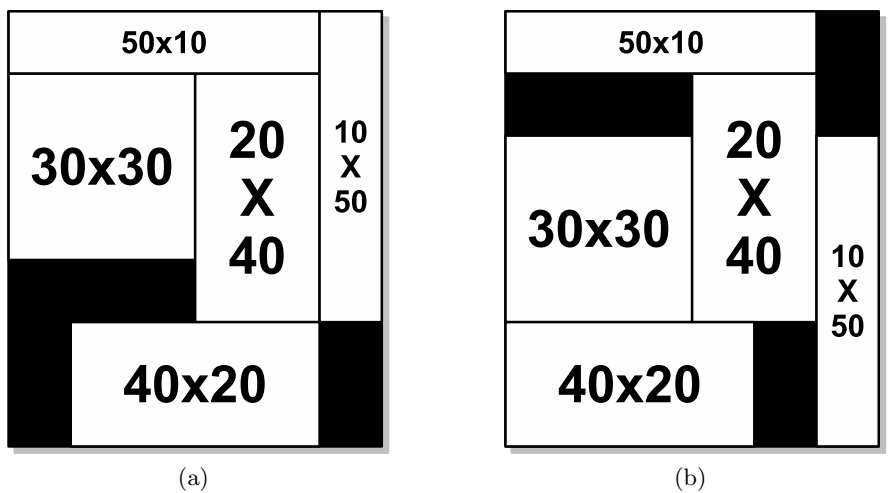

Figure 6: Examples of mapping solutions to one where rectangles are in their left-most, bottom-most positions.

## 5.2 Overall Strategy

Given an instance from our high-precision benchmark described in rational numbers, we multiply all values by the least common multiple of the denominators to get an instance with integer dimensions. We then apply the absolute placement solution techniques, with improvements we will subsequently explain, in order to find the optimal solutions. Once found, we divide all $x$- and $y$-coordinates describing the optimal solutions by the initial scaling constant in order to obtain the optimal solutions for the original problem.

Note that we can map every solution to one where all rectangles are slid over to the left and to the bottom as much as possible (Chazelle, 1983). For example, the solution in Figure 6a can be transformed into that of Figure 6b. Since all rectangles are now propped up from the left and below by other rectangles, each rectangle's $x$-coordinate is the sum of a subset of the widths of the other rectangles and each rectangle's $y$-coordinate is the sum of a subset of the heights of the other rectangles. Similarly, the width and height of the bounding box must be the sum of a subset of the widths and heights of the rectangles, respectively.

In the following subsections we first explain our techniques with respect to oriented instances, and then follow with how to handle the unoriented case.

## 5.3 Minimum Bounding Box Problem

Since we build the initial set of bounding boxes from all pairwise combinations of widths and heights within given ranges, the space is pruned by considering only bounding box widths and heights equal to the subset sums of the rectangle widths, and the subset sums of the rectangle heights, respectively. Recall from Section 4.4 that for every bounding box width, we compute a lower bound on the height. We further modify this by rounding the resulting bound up to the next subset sum of the rectangle heights.

### 5.3.1 Precomputing Subset Sums

We compute the set of all subset sums prior to searching. For oriented rectangles which cannot be rotated we compute two sets: one based only on the heights of the rectangles representing the candidate $y$-coordinates, and one based just on their widths representing the candidate $x$-coordinates. This distinction generates fewer subset sums compared to a single set of subset sums generated from both widths and heights.

### 5.3.2 Pruning Combinations of Widths and Heights

We can reject some bounding boxes for which certain width and height combinations are infeasible. This pruning technique relies on the observation that in certain cases, there may be only one unique set of rectangles that generate a specific width (height) for the bounding box.

For example, consider a bounding box width which can only be generated by a unique set of rectangles. Now assume that the heights of the same set of rectangles also uniquely determine the subset sum for a specific bounding box height. We say that this combination of bounding box width and height is incompatible. The reason is that this set of rectangles is the only way we can have a bounding box of the given width, and that implies this set of rectangles must appear in the solution laid out horizontally to one another. Thus, the same set of rectangles cannot appear stacked vertically in the solution. This contradicts the implications of a bounding box of the given height. Note that in this particular example, the only compatible height is the maximum height of the rectangles.

### 5.3.3 Learning From Infeasible Attempts

Recall that the algorithm for solving the minimal bounding box problem repeatedly calls the algorithm to solve the containment problem. Bounding boxes are tested in order of non-decreasing area until the first boxes with solutions are found. We can learn from the infeasible attempts.

For example, consider having to pack $N$ rectangles $\{r_1, r_2, ..., r_N\}$. Note that we use a pre-computed variable order for the rectangles. Let $r_d, d < N$ be the rectangle corresponding to the deepest in the search tree our depth-first search was able to go, during the entire search effort for a given bounding box. If the containmnet solver says this bounding box is infeasible, then the next bounding box height that we should consider can be the next greatest subset sum based on the smaller set $\{r_1, r_2, ..., r_{d+1}\}$ instead of considering all $N$ rectangles. The intuition behind this is that since our containment solver failed to even find an arrangement for the first $d + 1$ rectangles, it doesn't make sense to involve any of the remaining rectangles $\{r_{d+2}, ...r_N\}$ in the next largest subset sum for the bounding box height.

This method resembles conflict-directed backtracking. In our implementation, we also consider the effect of pruning using the wasted space heuristic as well.

## 5.4 Containment Problem

Similar to our low-precision methods, we first assign $x$-coordinates for the rectangles, then conduct a perfect packing transformation, and finally work on the $y$-coordinates (Huang

& Korf, 2010). The main difference between our high-precision methods and our low-precision methods are that instead of considering all possible integers as the domain of $x$- and $y$-coordinates, we consider the smaller set of subset sums of the widths and heights of the rectangles. The methods for using $x$-intervals remain unchanged and so we only describe how we search individual $x$-coordinates here.

### 5.4.1 ASSIGNING X-COORDINATES

For oriented rectangles, we choose $x$-coordinates from the set of subset sums of rectangle widths. Instead of precomputing the set as we did in the minimal bounding box problem, here we generate it dynamically at every node during the search prior to branching on various $x$-coordinate value assignments. The set is computed as follows:

1. Initialize the set with the value 0, as this represents placing a rectangle against the left side of the bounding box.

2. For every rectangle $r$ already assigned an $x$-coordinate at this point of the search, insert into the set the sum of its $x$-coordinate and its width. This represents placing a rectangle against the right side of $r$.

3. For every rectangle with its $x$-coordinate still unassigned, add its width to every element in our set, and insert the new sums back into the set.

### 5.4.2 PERFECT PACKING TRANSFORMATION

After assigning $x$-coordinates, we create a number of $1 \times 1$ rectangles to account for all empty space in the original instance. The transformation results in a new instance, with no empty space, and consists of the original rectangles plus the new $1 \times 1$ rectangles. Then for a given empty corner in a partial solution, we ask which of the original unplaced rectangles might fit there, or a $1 \times 1$ rectangle, essentially modeling empty corners as variables and rectangles as values.

In our high-precision benchmark, solving $N$=15 requires creating over 1.5 billion $1 \times 1$ rectangles because we scaled the problem up by a very large number. Here our packer simply requires too much memory and time. We avoid this problem by creating fewer and much larger rectangles to account for the empty space.

**Widening Existing Rectangles**  Assume in Figure 7a that the task is to pack three rectangles. Here we have a $10 \times 20$, $20 \times 10$, and a $40 \times 10$ rectangle in a $60 \times 50$ bounding box, and assume we have assigned $x$-coordinates but not $y$-coordinates. Given that the $x$-coordinates are already assigned, in any resulting packing solution the space to the right of the $40 \times 10$ rectangle must always be empty. Thus, we replace the $40 \times 10$ rectangle with a $60 \times 10$ rectangle, effectively widening the original rectangle. Likewise, we replace the $20 \times 10$ rectangle with a $30 \times 10$ rectangle, and the $10 \times 20$ rectangle by a $30 \times 20$ rectangle, as in Figure 7b. Our packer greedily attempts to widen the rectangles towards the right before widening them towards the left. After solving the problem we can just return the rectangles back to their original widths. This avoids creating many $1 \times 1$ rectangles during the perfect packing transformation to represent empty space.

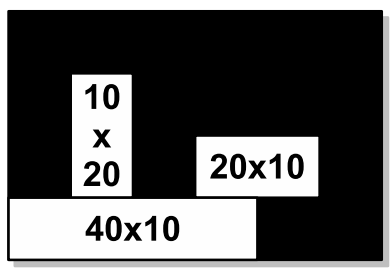

(a) A partial solution where only $x$-coordinates are known.

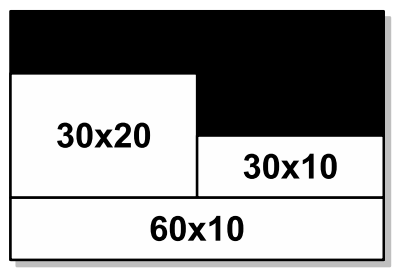

(b) The result of widening the rectangles.

Figure 7: Widening existing rectangles.

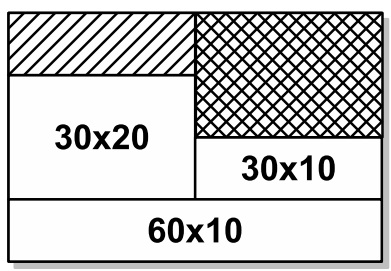

(a) A partial solution where only $x$-coordinates are known.

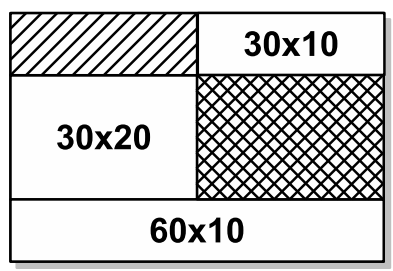

(b) A solution without $60 \times 1$ rectangles for empty space.

Figure 8: Consolidating empty space into horizontal strips.

**Turning Empty Space Into Large Rectangles**   In the partial solution of Figure 8a, we have assigned only the $x$-coordinates of the rectangles in a $60 \times 40$ bounding box. Instead of creating three hundred $1 \times 1$ rectangles to represent the empty space indicated by the single hash marks, we can use ten $30 \times 1$ rectangles without losing any packing solutions. Similarly, we represent the doubly-hashed empty space with twenty $30 \times 1$ rectangles instead of six hundred $1 \times 1$ rectangles. Note that we cannot use $60 \times 1$ rectangles for the empty space since we would inadvertently prune out the potential solution in Figure 8b.

### 5.4.3 Assigning Y-Coordinates

After the perfect packing transformation, we assign $y$-coordinates by asking which rectangle can be placed in a given empty corner. As before, we enforce the constraint that the $y$-coordinate of each rectangle must be a subset sum of the rectangle heights. Note that the rectangles we create via the perfect packing transformation are not included in the subset sum calculations, since they represent empty space.

### 5.4.4 Handling Unoriented Instances

For unoriented instances, when computing the initial bounding box widths and heights, we generate a single set of subset sums using both widths and heights from all rectangles in the instance instead of keeping the widths separated from the heights. Likewise, when generating the set of candidate $x$- and $y$-coordinates, we must add a fourth step to the bulleted list in subsection 5.4.1 where we add the height of every rectangle which has not yet been placed, to every element in the set of subset sums, as this represents the possibility of rotating the rectangle.

| Size $N$ | Optimal Solution | LCM | Bits of Precision | HK10 Boxes | Subsets Boxes | Mutex Boxes | HK11 Boxes |
|---|---|---|---|---|---|---|---|
| 1 | 1/2×1 | 2 | 2 | 1 | 1 | 1 | 1 |
| 2 | 1/2×4/3 | 6 | 6 | 1 | 1 | 1 | 1 |
| 3 | 1/2×19/12 | 12 | 8 | 2 | 2 | 2 | 2 |
| 4 | 5/6×1, 1/2×5/3 | 60 | 12 | 30 | 5 | 4 | 4 |
| 5 | 1/2×17/10 | 60 | 12 | 20 | 7 | 7 | 7 |
| 6 | 1/2×107/60 | 420 | 18 | 1,979 | 59 | 44 | 29 |
| 7 | 1/2×107/60 | 840 | 20 | 4,033 | 151 | 107 | 46 |
| 8 | 1/2×163/90 | 2,520 | 23 | 39,357 | 693 | 465 | 124 |
| 9 | 1/2×163/90 | 2,520 | 23 | 13,571 | 1,083 | 755 | 192 |
| 10 | 1/2×1817/990 | 27,720 | 30 | 2,682,948 | 7,489 | 4,901 | 585 |
| 11 | 1/2×7367/3960 | 27,720 | 30 | | 31,196 | 22,822 | 1,641 |
| 12 | 1/2×67/36 | 360,360 | 37 | | 66,425 | 38,827 | 2,366 |
| 13 | 1/2×185/99 | 360,360 | 37 | | 289,217 | 162,507 | 5,027 |
| 14 | 1/2×169/90 | 360,360 | 37 | | 549,135 | 382,059 | 9,548 |
| 15 | 1/2×79/42 | 720,720 | 39 | | 1,171,765 | 651,041 | 15,334 |

Table 10: The minimum-area bounding boxes and number of bounding boxes tested when packing unoriented rectangles $\frac{1}{1} \times \frac{1}{2}$, $\frac{1}{2} \times \frac{1}{3}$, $\frac{1}{3} \times \frac{1}{4}$, ..., and $\frac{1}{N} \times \frac{1}{N+1}$.

## 5.5 Experimental Results

We present two different data tables, one relating to improvements in the minimal bounding box problem measured by the number of bounding boxes tested, and another one on the overall CPU time for solving the entire rectangle-packing problem. We can separate our experiments this way because our solution schema decouples the minimal bounding box problem from the containment problem.

### 5.5.1 Minimum Bounding Box Problem

Table 10 shows the optimal solutions for our unoriented high-precision rectangle benchmark along with various properties of the corresponding instances. The first two columns give the problem size and the dimensions of the optimal solutions, respectively. The third gives the least common multiple of the first $N+1$ integers. The fourth is the number of bits required to represent the area of the minimal bounding box. Note that all but one of the optimal solutions have a width of $\frac{1}{2}$, since the first rectangle is much larger than any of the others. For $N=12$ and larger, the required precision exceeds that of a 32-bit integer.

The fifth through eighth columns compare the number of bounding boxes that various packers test to find all optimal solutions on our unoriented high-precision rectangle bench-mark. For each column going from left to right, we add one new technique for the minimal bounding box problem.

| Size $N$ | HK10 Time | Empty Space Time | Dynamic Time | HK11 Time |
|---|---|---|---|---|
| 6 | :00 | :00 | :00 | :00 |
| 7 | :02 | :00 | :00 | :00 |
| 8 | 1:11 | :00 | :00 | :00 |
| 9 | 1:51 | :03 | :00 | :00 |
| 10 | | 1:57 | :02 | :01 |
| 11 | | 41:40 | :57 | :18 |
| 12 | | 7:30:26 | 6:38 | :33 |
| 13 | | | 2:20:12 | 16:41 |
| 14 | | | 1:05:56:14 | 46:56 |
| 15 | | | | 4:28:20 |

Table 11: CPU times of various packers to find all minimum-area bounding boxes containing unoriented rectangles $\frac{1}{1} \times \frac{1}{2}$, $\frac{1}{2} \times \frac{1}{3}$, $\frac{1}{3} \times \frac{1}{4}$, ..., and $\frac{1}{N} \times \frac{1}{N+1}$.

HK10 is the number of bounding boxes required when simply scaling up the problem to an instance described completely in integers. The column called Subsets improves upon the second by testing only those bounding boxes whose dimensions are constrained by our subset sums technique. The column called Mutex improves upon the third by rejecting bounding boxes if the subset sum corresponding to its width is mutually exclusive to the subset sum corresponding to its height. HK11 improves upon the previous packer by using information learned from an infeasible attempt to reject future bounding boxes.

Using all improvements, by $N$=10 we test 4,500 times fewer bounding boxes compared to the previous state-of-the-art. On this instance HK10 ran out of memory on the last bounding box because of the sheer number of $1 \times 1$ rectangles created during the perfect packing transformation. The introduction of the prime number 11 as a denominator in the problem instance is responsible for the increased difficulty between $N$=9 and $N$=10.

### 5.5.2 Containment Problem

Table 11 compares the performance of various packers using our techniques. Because we have decoupled the minimal bounding box problem from the containment problem, in this table we use all of our optimizations for the minimal bounding box problem, and only compare the individual techniques applied to the containment problem. Therefore, the performance data reported is what is required to solve the overall problem using various containment problem packers.

The first column gives the size of the problem instance from our high-precision rectangle benchmark. As in previous tables, each successive column from left to right improves upon the previous column by an additional technique. The column called HK10 corresponds to using the previous state-of-the-art with our improved minimal bounding box algorithm. The column called Empty Space improves upon HK10 by precomputing all of the subset sums prior to searching for the $x$-coordinates, and uses our techniques to consolidate empty space in the $y$-coordinates. The column called Dynamic improves upon the previous one by dynamically computing subset sums. Finally, the last column called HK11 adds the ability to learn which unplaced rectangles to exclude from the subset sums computation after exploring an infeasible subtree. This data was collected using a Linux eight core 3GHz Intel Xeon X5460 without parallelization.

At $N$=10, the problem was scaled up 27,720 times in both dimensions, requiring HK10 to create 6,597,361 $1 \times 1$ units of empty space during the perfect packing transformation and causing it to run out of memory. Empty Space could not complete $N$=13 within a day because of the sheer number of subset sums that must be explored for both $x$- and $y$-coordinates, a problem avoided by Dynamic.

### 5.5.3 Comparison to Relative Placement

It is interesting to note that the number of bounding boxes appears to be increasing exponentially, mostly likely due to the exponential growth of the number of subset sums introduced by each successive rectangle in our high-precision benchmark. The difficulty of our unoriented high-precision rectangle benchmark is compounded by the fact that as the precision increases, the branching factor for the single $x$- and $y$-coordinate values in the containment problem also increases.

In contrast to our absolute placement technique, Moffitt and Pollack's (2006) relative placement techniques do not enumerate the different exact locations for the rectangles, and therefore promise to be immune to the problem of high-precision rectangles. They used a variable for every pair of rectangles to represent the relations above, below, left, and right. Their search algorithm then required at least one of these non-overlapping constraints to be true for every pair of rectangles. Their meta-CSP approach was modeled after work by Dechter, Meiri, and Pearl (1991) on solving binary constraint satisfaction problems, and included various pruning techniques such as model reduction, symmetry breaking, and graph-based pruning heuristics (Korf et al., 2010). They solve the minimum bounding box problem with a branch-and-bound algorithm, evaluating the size of the bounding box when all non-overlapping relationships have been determined, and keeping track of the bounding box of smallest area seen so far.

Note that by contrast, our solver tests bounding boxes in order of non-decreasing area. Also, the size of their formulation uses $N^2$ variables while we use only $N$. Finally, their packer only returns one optimal solution as opposed to ours, which does more work by returning all of the optimal solutions.

We have been able to benchmark their code on our machine in order to provide some kind of comparison between their methods and ours. This is a crude comparison, because we cannot run their packer on our unoriented high-precision rectangle benchmark since they have hard-coded into their packer the unoriented consecutive-rectangle benchmark, a much easier benchmark as we have shown in Table 7.

The first column in Table 12 refers to the problem size. The second column called MP06 gives the CPU time required for Moffitt and Pollack's code on problem instances from the unoriented consecutive-rectangle benchmark, which uses low-precision rectangles. The third column called HK11 gives the CPU time required by our packer on problem instances from the unoriented high-precision rectangle benchmark. Each data point in this table was collected using an eight core 3GHz Intel Xeon X5460 in Linux without parallelization. Note that our algorithm packs the same number of rectangles somewhat faster than that of Moffit and Pollack's.

| Size $N$ | MP06 Time | HK11 Time |
|:---:|:---:|:---:|
| 10 | :03 | :01 |
| 11 | :13 | :18 |
| 12 | 2:26 | :33 |
| 13 | 17:40 | 16:41 |
| 14 | 1:48:09 | 46:56 |
| 15 | 7:27:42 | 4:28:20 |

Table 12: CPU times required by Moffitt and Pollack's packer on the unoriented consecutive-rectangle and our packer on the unoriented high-precision rectangle benchmarks.

### 5.6 Summary of High-Precision Rectangles

In this section we proposed a new benchmark consisting of instances with rectangles of high-precision dimensions as well as techniques for using subset sums to limit the number of positions that must be considered, rules to filter out subset sums for both the minimal bounding box and containment problems, methods to learn from infeasible subtrees, and ways to reduce the number of rectangles created during the perfect packing transformation. These techniques exploit no special properties of the benchmark, but are most useful for rectangles with high-precision dimensions.

Using all of our methods, we solved six more problems up to $N=15$ on our new benchmark compared to using our low-precision packer on a scaled up instance. Our packer is over two orders of magnitude faster at $N=9$ than the previous state-of-the-art, and tests 4,500 times fewer bounding boxes. A cursory comparison between the state-of-the-art using the relative placement search space and our own shows that we perform slightly faster than Moffitt and Pollack's packer, on a benchmark which we have previously shown in Section 4.3.2 to be significantly more difficult than the unoriented consecutive-rectangle benchmark that Moffitt and Pollack's program was run on.

## 6. Future Work

Humans solve jigsaw puzzles both by asking where a particular piece should go, as well as asking what piece should go in some empty region. Our packer makes use of both models, the former for the $x$-coordinates and the latter for the $y$-coordinates. It would be interesting to see how applicable this dual formulation is in other packing, layout, and scheduling problems. Currently, we work on the $x$-coordinates by asking "Where does this go?", and we work on the $y$-coordinates by asking "What goes in this location?" Our method has reduced the time spent in the $y$-coordinates so much that now the time spent working on the $x$-coordinates is orders of magnitude greater than the time spent working on the $y$-coordinates. This suggests that performance might be improved by considering both models simultaneously.

As another direction for continued work, the data indicates that the number of bounding boxes explored by our minimum bounding boxes solver is the main bottleneck to solving larger instances of our unoriented high-precision rectangle benchmark. An observation we can make is that across many of these bounding boxes, the same partial solutions are being explored, resulting in much redundant computation. Consequently, a branch-and-bound method that starts with a large bounding box, and gradually reduces its dimensions while various packings are explored would be a promising avenue of further research.

## 7. Conclusions

We have presented several new improvements to the previous state-of-the-art in optimal rectangle packing. Within the schema of assigning $x$-coordinates prior to $y$-coordinates, we introduced a dynamic variable order for the $x$-coordinates, and a constraint that adapts Korf's (2003) wasted space pruning heuristic to the one-dimensional case. For the $y$-coordinates we work on the perfect packing transformation of the original problem, by using a model that assigns rectangles to empty corners, and inference rules to reduce the model's variables.

Our improvements in the search for $y$-coordinates helped us solve $N$=27 of the consecutive-square benchmark over an order of magnitude faster than the previous state-of-the-art, and our improvements in the search for $x$-coordinates also gave us another order of magnitude speedup by $N$=28, compared to leaving those optimizations out. With all our techniques, we are over 19 times faster than the previous state-of-the-art on the largest problem solved to date, allowing us to extend the known solutions for the consecutive-square benchmark from $N$=27 to $N$=32. Furthermore, the data show that very little time is spent searching for $y$-coordinates, suggesting that rectangle packing may be largely reduced to the problem of determining the x-coordinates.

All of the techniques presented to pick $y$-coordinates are tightly coupled with the dual view of asking what must go in an empty location. Furthermore, while searching for $x$-coordinates, our pruning rule is based on the analysis of irregular regions of empty space, and our dynamic variable order also rests on the observation that less empty space leads to a more constrained problem. The success of these techniques in rectangle packing make them worth exploring in many other packing, layout, and scheduling problems.

We have also introduced two new benchmarks, one oriented and one unoriented, that include rectangles of various aspect ratios. These new benchmarks avoid various properties of easy instances, which we have identified, and were shown to be much harder through a side-by-side comparison between various benchmarks using the same state-of-the-art packer. We have also proposed several search strategies to improve performance on our new benchmarks. We improved upon our strategies used to handle dominance conditions, proposed a variable ordering heuristic based on increasing branching factor that generalizes previous strategies, tuned a global interval parameter, and introduced a method to balance the sizes of the intervals assigned to the $x$-coordinate variables.

Our experiments revealed that it takes orders of magnitude more time to solve our new benchmarks compared to instances from the consecutive-square benchmark with the same number of rectangles. We therefore advocate the inclusion of these new, more difficult benchmarks in the suite of benchmarks used for research in optimal rectangle packing. Finally, using all of our techniques together, we solved $N=21$ of the oriented equal-perimeter benchmark about 500 times faster, and $N=16$ of the unoriented double-perimeter benchmark about 40 times faster than simply using methods tuned for consecutive-squares.

In order to test the limits of our rectangle packer, we presented a new high-precision benchmark specifically capturing the pathological case where each successive rectangle quickly increases the precision required to represent coordinate locations. We presented various techniques to adapt the absolute placement approach to handle these types of instances, including dynamically using subset sums to limit the number of coordinate values that must be tested, mutex reasoning that allows us to reject certain combinations of subset sums used for a bounding box's width and height, a general method for rejecting future subset sums based on a previously infeasible search, and finally a memory-efficient adaptation of our perfect packing transformation to high-precision rectangle instances.

We solved $N=12$ of the high-precision benchmark in half a minute, 800 times faster than a basic version of our packer augmented with only the high-precision version of our perfect packing inference rules so that it did not run out of memory. This was also the first instance requiring precision exceeding the capacity of a 32-bit integer. Our techniques allowed us to solve up to $N=15$ compared to $N=9$, the largest instance our low-precision techniques alone could solve. Our methods also reduced the number of bounding boxes generated by a factor of 4,500. At this point we are solving problems that require a minimum of 39 bits of precision, which should meet the requirements of many real-world problems.

We then provided a comparison to the state-of-the-art relative placement packer showing that our absolute-placement packer remains competitive even on rectangles of high-precision, and reported on promising avenues of research which may potentially give the absolute placement approach a clear competitive edge over relative-placement methods.

Although we have mainly focused on obtaining optimal solutions in our benchmarks, our work may be easily adapted to applications requiring quick suboptimal solutions by simply replacing our algorithm for the minimum bounding box problem with alternatives such as the anytime algorithm that we described in Section 3.3.1.

## 7.1 Comparison to Constraint Programming Methodologies

There are clearly tradeoffs between taking our ground-up programming approach in C++ and taking a constraint programming approach. While the latter provides quick prototyping and reuse of constraint libraries that other researchers have already implemented, it also forces the problem to be expressed in the abstract constraint language. Such an abstract layer turns out to add unnecessary overhead for the algorithms and data structures that one naturally uses to solve our problem of optimal rectangle packing.

For example, as we previously described, for the cumulative constraint, we simply add a constant to a consecutive range in an integer array when we assign an $x$-coordinate to a rectangle. When we backtrack, we scan the same array and just subtract the same constant. Scanning and manipulating arrays, iteration, and fast pushing and popping of the program stack in recursive algorithms are precisely the operations that modern computer hardware has been optimized for. This is significant as we explore over two trillion search nodes for $N$=32 in the square-packing benchmark, and in fact our solver spends about 75% of its time on just these array manipulation operations alone. This is how we explain the orders of magnitude speedup for processing just the $x$-coordinate solutions in a 1D array instead of the 2D bitmap by Korf (2003). As we move from 1D arrays, to 2D bitmaps, to abstract representations of variables and values in constraint programming, the patterns of computation and data structures simply become too distant from what the underlying hardware is optimized for.

For optimal rectangle packing, it happens that the algorithms and data structures that naturally solve the problem map very nicely in form and function to the hardware of modern computers. Note that one may always port this code into a constraint module that may be called by a constraint solver, but there is still some computational indirection between this module and the backtracking control logic of the constraint solver. The sacrifice we make in our approach, however, is the fact that our solver is tailored specifically to the rectangle packing problem as we have defined it, and it would require more implementation effort to reconfigure our algorithms and heuristics for a slightly different rectangle packing problem. We hope, however, that this latter problem is ameliorated by disciplined object-oriented, modular software design.

## 8. Broader Lessons

Beyond the specific problem of rectangle packing, what broader lessons can we learn from this work? We believe there are several.

One of the main applications of rectangle packing is to scheduling. As described in the introduction, the rectangle packing problem is an abstraction of a scheduling problem where different tasks take different amounts of time, and all require different amounts of a one-dimensional resource that must be allocated contiguously, such as memory on a computer. The width of the bounding box becomes the total time, the height the total amount of the resource available, and each job becomes a rectangle with width equal to time duration, and height equal to the amount of the resource required.

What we found, however, is that vast majority of the time used by our rectangle packer is in assigning just the $x$-coordinates of the rectangles, subject to the cumulative constraint, which is that for every $x$-coordinate in the bounding box, the sum of the heights of the rectangles that overlap that $x$-coordinate cannot exceed the height of the bounding box. This important subpart of the rectangle-packing problem models a much more general problem known as the resource-constrained scheduling problem. This is the same as the scheduling problem described above, but without the constraint that the resource be allocated contiguously. For example, in scheduling tasks on a planetary rover with a limited power budget, the sum of the power requirements of all the tasks that are active at any given time cannot exceed the total power budget of the rover. Thus, this subpart of our rectangle packer can be used to tackle this more general scheduling problem.

Another general lesson that can be learned from this work is that the absolute placement approach to various packing problems in two, three, or more dimensions may be effective even on problems with high precision dimensions. One might expect that absolute placement would not be competitive with relative placement approaches on these problems, but the key to our success in this area is that instead of considering all possible placements, we only consider placements that correspond to subset sums of the relevant dimensions. While there is no guarantee that this approach will work in other high-precision packing problems, we have shown that it is at least worth considering, and may be effective.

Perhaps the largest lesson to be learned here is both encouraging and discouraging. The problem of rectangle packing is extremely simple, and can be understood by and played as a game by children. Yet the research over the last decade described here shows that the most efficient algorithms are quite complex. If the best algorithms for such a simple problem are so complex, it is likely that the best algorithms for more complex problems are even more complex, which is the discouraging part. The encouraging part is that the history of this research has shown that each new idea can result in an order of magnitude improvement over the previous state of the art on larger problems, suggesting that there is still very significant progress to be made on this problem, and by extension others like it.

## Acknowledgments

We wish to thank Reza Ahmadi, Adnan Darwiche, and Adam Meyerson for their advice on this work. We also thank Michael Moffitt for making his packer available. This research was funded in part by the National Science Foundation under grant number IIS-0713178. The source code of our optimal rectangle packer is open sourced and available at `http://code.google.com/p/rectpack`.